\documentclass{article}

\usepackage{microtype}
\usepackage{graphicx}
\usepackage{subfigure}
\usepackage{booktabs} %

\usepackage{hyperref}

\usepackage[accepted]{icml2025}

\usepackage{amsmath}
\usepackage{amssymb}
\usepackage{mathtools}
\usepackage{amsthm}

\usepackage[capitalize,noabbrev]{cleveref}

\theoremstyle{plain}

\theoremstyle{definition}

\theoremstyle{remark}

\usepackage[textsize=tiny]{todonotes}

\usepackage{xspace}
\newcommand{\method}{\textsc{CpSDE}\xspace}
\newcommand{\dockmodel}{\textsc{AtomSDE}\xspace}
\newcommand{\seqmodel}{\textsc{ResRouter}\xspace}
\usepackage{dsfont} %
\usepackage{multirow} %
\usepackage{adjustbox} %
\usepackage{paralist} %
\usepackage{longtable, lscape}
\usepackage{colortbl} %
\definecolor{myred}{RGB}{255, 210, 210}
\definecolor{mygreen}{RGB}{210, 255, 210}

\usepackage{bbding}
\usepackage{pifont}

\usepackage{algorithm}
\usepackage{algorithmic}
\usepackage{tikz}
\usepackage{wrapfig}

\usepackage[compact]{titlesec}
\usepackage{sidecap}
\titlespacing*{\section}{0pt}{*1.0}{*0.8}
\titlespacing*{\subsection}{0pt}{*0.5}{*0.3}
\titlespacing*{\subsubsection}{0pt}{*0.3}{*0.3}
\allowdisplaybreaks

\icmltitlerunning{Designing Cyclic Peptides via Harmonic SDE with Atom-Bond Modeling}

\usepackage{amsmath,amsfonts,bm}

\def\eqref#1{equation~\ref{#1}}

\def\1{\bm{1}}

\def\rvb{{\mathbf{b}}}

\def\rve{{\mathbf{e}}}
\def\rvf{{\mathbf{f}}}

\def\rvh{{\mathbf{h}}}

\def\rvs{{\mathbf{s}}}
\def\rvt{{\mathbf{t}}}

\def\rvw{{\mathbf{w}}}
\def\rvx{{\mathbf{x}}}

\def\rmA{{\mathbf{A}}}

\def\rmD{{\mathbf{D}}}

\def\rmH{{\mathbf{H}}}
\def\rmI{{\mathbf{I}}}

\def\rmL{{\mathbf{L}}}

\def\rmP{{\mathbf{P}}}

\def\rmT{{\mathbf{T}}}

\def\rmX{{\mathbf{X}}}

\def\vtheta{{\bm{\theta}}}

\def\mD{{\bm{D}}}

\def\mLambda{{\bm{\Lambda}}}

\DeclareMathAlphabet{\mathsfit}{\encodingdefault}{\sfdefault}{m}{sl}
\SetMathAlphabet{\mathsfit}{bold}{\encodingdefault}{\sfdefault}{bx}{n}

\def\gA{{\mathcal{A}}}

\def\gC{{\mathcal{C}}}

\def\gG{{\mathcal{G}}}

\def\gL{{\mathcal{L}}}

\def\gN{{\mathcal{N}}}

\def\gP{{\mathcal{P}}}

\def\gT{{\mathcal{T}}}
\def\gU{{\mathcal{U}}}

\newcommand{\R}{\mathbb{R}}

\newcommand{\ud}{\mathrm{d}}

\begin{document}

\twocolumn[
\icmltitle{Designing Cyclic Peptides via Harmonic SDE with Atom-Bond Modeling}

\icmlsetsymbol{equal}{*}

\begin{icmlauthorlist}
\icmlauthor{Xiangxin Zhou}{equal,seed,ucas,casia}
\icmlauthor{Mingyu Li}{equal,thuair,sjtu}
\icmlauthor{Yi Xiao}{thuair}
\icmlauthor{Jiahan Li}{thuair}
\icmlauthor{Dongyu Xue}{seed}
\icmlauthor{Zaixiang Zheng}{seed}
\newline
\icmlauthor{Jianzhu Ma}{thuair,thuee}
\icmlauthor{Quanquan Gu}{seed}
\end{icmlauthorlist}

\icmlaffiliation{seed}{ByteDance Seed~(Work was done during Xiangxin's internship at ByteDance Seed.)}
\icmlaffiliation{ucas}{School of Artificial Intelligence, University of Chinese Academy of Sciences}
\icmlaffiliation{casia}{New Laboratory of Pattern Recognition (NLPR), State Key Laboratory of Multimodal Artificial Intelligence Systems (MAIS), Institute of Automation, Chinese Academy of Sciences (CASIA)}
\icmlaffiliation{thuair}{Institute for AI Industry Research, Tsinghua University}
\icmlaffiliation{sjtu}{School of Medicine, Shanghai Jiao Tong University}
\icmlaffiliation{thuee}{Department of Electronic Engineering, Tsinghua University}

\icmlcorrespondingauthor{Quanquan Gu}{quanquan.gu@bytedance.com}

\icmlkeywords{Machine Learning, ICML}

\vskip 0.3in
]

\printAffiliationsAndNotice{\icmlEqualContribution} %

\begin{abstract}

Cyclic peptides offer inherent advantages in pharmaceuticals. For example, cyclic peptides are more resistant to enzymatic hydrolysis compared to linear peptides and usually exhibit excellent stability and affinity. Although deep generative models have achieved great success in linear peptide design, several challenges prevent the development of computational methods for designing diverse types of cyclic peptides. These challenges include the scarcity of 3D structural data on target proteins and associated cyclic peptide ligands, the geometric constraints that cyclization imposes, and the involvement of non-canonical amino acids in cyclization. To address the above challenges, we introduce \method, which consists of two key components: \dockmodel, a generative structure prediction model based on harmonic SDE, and \seqmodel, a residue type predictor. Utilizing a routed sampling algorithm that alternates between these two models to iteratively update sequences and structures, \method facilitates the generation of cyclic peptides. By employing explicit all-atom and bond modeling, \method overcomes existing data limitations and is proficient in designing a wide variety of cyclic peptides.
Our experimental results demonstrate that the cyclic peptides designed by our method exhibit reliable stability and affinity.

\end{abstract}

\section{Introduction}

Therapeutic peptides are a distinct group of pharmaceutical compounds comprising a sequence of precisely arranged amino acids \citep{driggers2008exploration,tsomaia2015peptide,zorzi2017cyclicthe}. 
Peptide drugs tend to exhibit lower toxicity, enhanced biological activity, and high target specificity compared to small molecules, superior cellular permeability, lower cost, lower immunogenicity compared to antibody drugs \citep{driggers2008exploration}. 
However, conventional linear peptides suffer from a short half-life, limited stability, and susceptibility to hydrolase degradation \citep{tsomaia2015peptide}, restricting their therapeutic potential and broader use.
\begin{figure}[t!]
\begin{center}
\centerline{\includegraphics[width=\columnwidth]{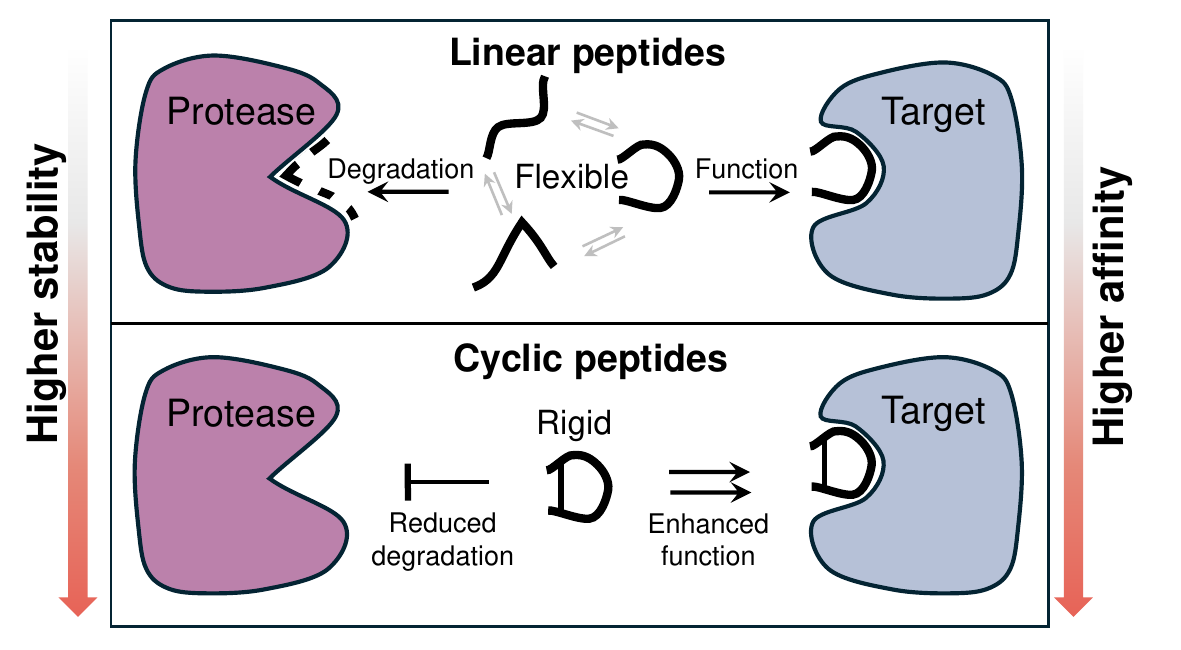}}
\caption{Comparative advantages of cyclic peptides over linear peptides. 
Linear peptides are easily degraded, whereas cyclic peptides are protected against enzyme hydrolysis, allowing them to function more effectively within the human body. Cyclic peptides generally exhibit better stability and affinity.}
\label{fig:motivation}
\end{center}
\end{figure}
Unlike linear peptides, as shown in \cref{fig:motivation}, cyclic peptides are chains of residues that typically form one or two closed loops, often incorporating non-natural residues. For example, a single loop can be created by connecting the N- and C-termini with a peptide bond or linking two internal cysteines through a disulfide bond \citep{camarero1999biosynthesis,giordanetto2014macrocyclic,kale2018cyclization}. Such cyclization enhances their resistance to digestive enzymes and enables them to bind protein surfaces with high affinity in more stable conformations. Traditional methods for discovering cyclic peptides involve chemical synthesis and high-throughput screening, both of which are labor-intensive and costly. 
This has led to adopting \textit{in silico} approaches such as virtual screening \citep{zotchev2006rational} and de novo design \citep{kawamura2017highly,peacock2021discovery,bhardwaj2022accurate,garcia2023macrocycles} to streamline the discovery process.

Cyclic peptides can adopt various structural forms depending on how their amino acid residues are linked together based on different chemical bonds and geometric distances, for example, the distance between two cysteines forming a disulfide bond typically falls from 2.0 to 2.5 \AA\ \citep{fass2012disulfide}. These structures are classified into four categories (see \cref{fig:overview}) based on the atoms of residues that form the cyclic structure: (1) \textbf{Head-to-tail cyclization.} The N-terminus (head) of one amino acid forms a peptide bond with the C-terminus (tail) of another amino acid, resulting in a closed ring. (2) \textbf{Side-to-tail cyclization.} The side chain of an amino acid is linked to the C-terminus (tail) of the peptide. (3) \textbf{Head-to-side cyclization.} The N-terminus (head) of one amino acid is linked to the side chain of another amino acid. (4) \textbf{Side-to-side cyclization.} The side chain of an amino acid is linked to the side chain of another, forming a cyclic structure that does not involve the head or tail of the peptide backbone. Therefore, incorporating specific chemical and geometric constraints (such as bond lengths, angles, and atom compositions) relating to one or more sets of residues is essential for designing different kinds of cyclic peptides.
Recent attempts have tried designing disulfide-linked cyclic peptides based on a post-processing strategy \citep{wang2024target} or head-to-tail cyclic peptides 
\citep{rettie2024accurate} using modified position encoding in protein generative models. However, these methods only consider one specific type of cyclic peptides and do not support other types based on specific constraints. Furthermore, the availability of real-world 3D structural data for protein-ligand complexes involving cyclic peptides is limited, posing a challenge to advancing computational cyclic peptide design.

To tackle these challenges, we developed the \method, which comprises two models: a harmonic-SDE-based generative structure prediction model named \dockmodel and a residue type predictor named \seqmodel. Unlike leading works \citep{watson2023novo,yim2023se} that typically use the residue frame representation for protein design, we employ the all-atom and bond representation. Since both linear and cyclic peptides are composed of atoms and bonds, this representation allows us to model interactions at the most fundamental level. It maximizes the use of small molecule and linear peptide data while minimizing reliance on cyclic peptide data. The inclusion of bond modeling effectively addresses the geometric constraints introduced by cyclization. With our designed routed sampling method, we iteratively update both the sequence and structure by alternating between the two models, which enables the generation of all types of cyclic peptides. We highlight our main contributions as follows:
\begin{compactitem}
\item We introduce \method, the first generative algorithm, to our knowledge, capable of directly generating all types of cyclic peptides informed by the 3D structure of a protein target, paving the way for developments in peptide-based drug discovery.
\item Our approach designs cyclic peptides with robust stability and affinity while maintaining high diversity, underscoring its significant potential in drug development and therapeutic innovation.
\item Through case studies involving molecular dynamics simulations, we demonstrate our method's practical utility and effectiveness in drug design, reinforcing its applicability and impact in real-world scenarios.
\end{compactitem}

\section{Related Work}
\label{sec:related_work}

\textbf{All-Atom Protein Design.}
Protein design traditionally involves first designing the backbone, followed by sequence design. Diffusion models \citep{ho2020denoising,song2021scorebased} have been applied to protein backbone design \citep{watson2023novo}. This process is typically followed by inverse-folding models \citep{dauparas2022robust}, enabling comprehensive protein design.
Recently, there has been a shift towards \textit{co-design}, where both protein sequences and structures are generated jointly \citep{jin2022iterative,luo2022antigen,kong2023end,lisanza2024multistate,campbell2024generative}.

However, all-atom\footnote{In some contexts, ``all'' in ``all-atom'' is interpreted to encompass all types of biomolecules, including small molecules, proteins, and nucleic acids \citep{krishna2024generalized}. In our work, however, we use ``all-atom'' with the same meaning as ``full-atom'', specifically to represent all heavy atoms within proteins.} structure modeling is essential for comprehending protein functionality, such as protein-protein interactions. Consequently, recent research has increasingly focused on all-atom protein design to gain a more detailed and accurate understanding.
To achieve full-atom antibody design,
\citet{kong2024fullatom} proposed to use a multi-channel equivariant layer to encode all-atom structures, and
\citet{martinkus2023abdiffuser} introduced a backbone and internal generic side chain representation. 
\citet{chen2025apm} adopted a representation that includes amino
acid type, backbone structure, and sidechain torsion angles for designing protein complexes.
A notable advance in \textit{de novo} all-atom protein design is Protpardelle~\citep{chu2024all}, which suggested modeling a ``superposition'' over possible side-chain states and introduced the atom73 representation. This inspires us to apply all-atom structures in a similar manner for \textit{de novo} cyclic peptide design. Unlike Protpardelle, we also incorporate bond modeling, which is crucial for ensuring successful cyclization.

\textbf{Peptide Design.}
Peptides, consisting of short chains of amino acid residues, are essential in numerous biological processes due to their interactions with various target molecules, offering substantial potential in drug discovery, such as targeting undruggable proteins \citep{hosseinzadeh2021anchor}. 
Traditional computational peptide design methods often rely on searching and sampling residues or motifs from chemical databases \citep{bhardwaj2016accurate}, which can be time-consuming and limit the diversity of designed structures \citep{cao2022design}. 
In contrast, deep generative models, known for their strong capability in modeling data distributions, have been applied to peptide design and have shown great potential. 
Several studies have explored designing peptide backbones \citep{boom2024scorebasedgenerativemodelsdesigning}, designing peptide sequences \citep{chen2024pepmlm}, or generating specific peptide structures such as $\alpha$-helices \citep{xie2023helixgan,xie2024helixdiff}. \citet{lin2024ppflow} proposed target-aware peptide sequence-structure co-design with flow matching on peptide global translation, orientation, backbone torsions, and sequences. 
Recent advances in full-atom protein design have significantly improved peptide generation capabilities.
For example, \citet{kong2024fullatom} employed a latent diffusion model on a latent space that encodes full-atom peptide structures. 
\citet{li2024full} proposed to represent peptides with backbone atoms and side-chain torsion angles, employing flow-based models to generate full-atom peptides given the protein targets. 
These methods face challenges in adapting to cyclic peptide design because they model protein structures at the residue level, which complicates the incorporation of covalent bonds or non-canonical amino acids essential for cyclization.

\section{Method}

\begin{figure*}[t!]
\begin{center}
\centerline{\includegraphics[width=0.98\textwidth]{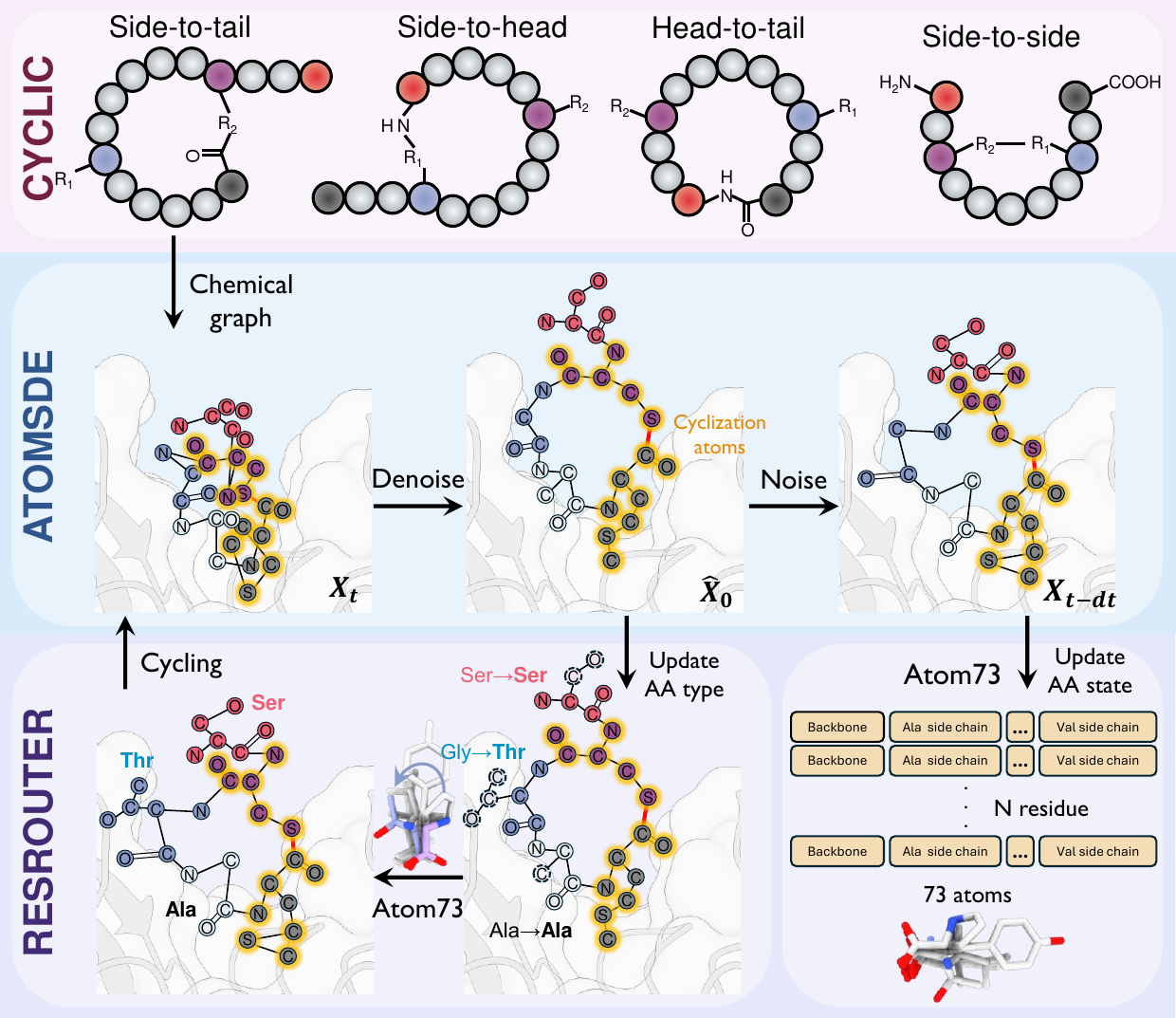}}
\vspace{-4mm}
\caption{Overview of \method. The generative process is structured as follows: (1) A cyclization type is initially selected, which subsequently determines the associated 2D chemical graph; (2) At time $t$, given the entire chemical graph defined by both cyclization (highlighted with a yellow shadow) and the predicted residue types, the all-atom structure is initially denoised using \dockmodel and then re-noised in accordance with the integration step of the reverse-time SDE. The updated structures are then preserved in the Atom73 representation; (3) \seqmodel predicts the residue types not constrained by cyclization, based on the denoised structure. Consequently, the chemical graph and the all-atom structures are updated using the Atom73 representation
Steps (2) and (3) are iteratively executed. The incorporation of the cyclization chemical graph, with chemical bonds as edges, ensures that the generated peptide forms a cyclic structure.}
\label{fig:overview}
\vspace{-10mm}
\end{center}
\end{figure*}

In this section, we present \method, a groundbreaking approach for designing cyclic peptides. It features an SDE-based generative structure prediction model, \dockmodel, and a residue type predictor, \seqmodel, both utilizing all-atom and bond modeling. We begin by defining the cyclic peptide design task and providing an overview of SDE-based generative models in \cref{subsec:preliminaries}. In \cref{subsec:dockmodel}, we detail \dockmodel, based on a harmonic SDE, and in \cref{subsec:seqmodel}, we describe \seqmodel, which predicts residue types based on denoised structures. Lastly, in \cref{subsec:sampling}, we explain how to alternate between these models through routed sampling to generate cyclic peptides.

\subsection{Preliminaries}
\label{subsec:preliminaries}

A peptide is a specific type of protein, generally composed of fewer than 30 amino acid residues.
The type of the $i$-th residue $a_i\in \{1,2,\ldots,20\}$ is determined by its side-chain R group. 
Thus, the \textbf{all-atom} 3D structure of a peptide specifies both its sequence and structure, inspiring us to focus on generating 3D coordinates of all atoms for peptide design.
Cyclic peptides, particularly their cyclization regions, often include unique inter-residue chemical bonds and occasionally non-canonical amino acids. Despite their non-canonical nature, these cyclization regions usually display distinct patterns. Additionally, providing detailed cyclization information during cyclic peptide design is essential, as it ensures wet-lab synthesizability. Here we denote the all-atom 3D structure of the cyclic peptide as $\gP$ and the 3D structures of the receptor (i.e., protein target) as $\gT$. We define the chemical graph of the cyclization parts as $\gC$, which contains atoms as nodes and chemical bonds as edges. Note that since the 3D structures of the cyclization part are unavailable, there is no information about atom positions in $\gC$. Our final goal is to model the conditional distribution $P(\gP|\gT,\gC)$, i.e., generate the cyclic peptides given the 3D receptor structure and the cyclization chemical graph.

We provide basic knowledge on stochastic differential equations (SDE) and SDE-based generative models \citep{song2019generative,ho2020denoising,song2021scorebased}.
SDE-based generative models (also known as score-based generative models and diffusion models) learn the data distribution by learning to denoise. The forward SDE injects noise gradually into the data $\rvx_0\in \R^d$ and constructs a diffusion process $\{\rvx_t\}_{t\in[0,1]}$, such that $\rvx_0\sim p_0$ and $\rvx_1\sim p_1$, where $p_0$ is the data distribution and $p_1$ is the prior distribution:
\begin{align}
    \ud\rvx = \rvf(\rvx,t)\,\ud t+ g(t)\ \ud \rvw,
\end{align}
where $\rvf(\cdot,t):\R^d\to\R^d$ is a vector-valued function known as drift coefficient, $g(\cdot): \R\to\R$ is a scalar function known as diffusion coefficient, and $\rvw$ is the standard Wiener process (also known as Brownian motion). The induced perturbation kernel $p_{0t}(\rvx_t|\rvx_0)$ is a Gaussian distribution that can be efficiently sampled.
The resultant $p_1(\rvx_t|\rvx_0)$ typically approximates a Gaussian distribution $p_1(\rvx_t)$ (i.e., prior distribution), which is independent of $\rvx_0$.

The reverse SDE starts from a sample from the prior distribution, denoises the noisy sample iteratively, and finally produces a generated sample:
\begin{align}
    \ud\rvx = [\rvf(\rvx,t) - g(t)^2 \nabla_\rvx \log p_t(\rvx)]\, \ud t + g(t) \, \ud\Bar{\rvw},
    \label{eq:reverse_sde}
\end{align}
where $\Bar{\rvw}$ is a standard Wiener process when time flows backwards from $1$ to $0$ and and $\ud t$ is an infinitesimal negative timestep.
Typically, a neural network $\rvs_\vtheta(\rvx_t,t)$ is used to approximate the underlying score function $\nabla_\rvx \log p_t(\rvx)$, and it can be trained via the following score matching objective:
\begin{align}
    \gL = \mathbb{E}_t[\lambda(t)\mathbb{E}_{\rvx_0} \mathbb{E}_{\rvx_t|\rvx_0}\Vert 
    \rvs_\vtheta(\rvx_t,t) 
    -
    \nabla_{\rvx_t} \log p_{0t}(\rvx_t|\rvx_0)
    \Vert^2],
    \nonumber
\end{align}
where $\lambda(t)$ is time-dependent weighting function and $t$ is uniformly sampled over $[0,1]$.

\subsection{\dockmodel}
\label{subsec:dockmodel}
To accurately model both atomic interactions and bond constraints, we initially train a docking model named \dockmodel using protein-ligand complex data, incorporating both small molecules and peptides as ligands.

In this subsection, for brevity, given a protein-ligand complex, we assume there are $N_L$ atoms whose coordinates are $\rvx^L \in \R^{N_L\times 3}$ in the ligand and $N_P$ atoms whose coordinates are $\rvx^P \in \R^{N_P\times 3}$ in the protein $\gT$. We denote the chemical graph of the ligand as $\gG_C$, where nodes are atoms and edges are chemical bonds.
We build a generative docking model \dockmodel to learn the conditional distribution $p(\rvx^{L}|\gT,\gG_C)$. 

We choose Variance Preserving (VP) SDE \citep{ho2020denoising} instead of Variance Exploding (VE) SDE \citep{song2019generative} for our scenario. The reason is that, when $t$ is large, VE SDE causes the noisy ligands to drift far from the receptor and introduces different spatial scales between the ligands and receptor, leading to a loss of valid interaction with the receptor. 
Inspired by \citet{jing2023eigenfold,stark2023harmonic}, we introduce a harmonic SDE to fully leverage the connection information embedded in the chemical graph. We define $\rmH \coloneqq\rmL+\sigma^{-2}_P\rmI$, where $\rmL=\rmD-\rmA$ is the Laplacian matrix of chemical graph $\gG_C$, $\rmD$ is the degree matrix, $\rmA$ is the adjacent matrix, and $\sigma_P$ is a receptor-dependent scalar value. The positive definite matrix can be decomposed as $\rmH=\rmP\mLambda\rmP^\intercal$ where $\rmP$ is an orthogonal matrix (i.e., $\rmP\rmP^\intercal=\rmI$) and $\mLambda=\text{diag}(\lambda_1,\ldots,\lambda_{N_L})$ is a diagonal matrix that contains the eigenvalues. 
We define the $\gG_C$-dependent forward SDE as follows:
\begin{align}
    \ud\rvx^L = -\frac{1}{2}\beta(t)\Tilde{\rvx}^L \, \ud t + \sqrt{\beta(t)}{\mLambda}^{\frac{1}{2}}\rmP^\intercal\,\ud \rvw,
    \label{eq:forward_harmonic_sde}
\end{align}
where $\beta(t)$ is a positive time-dependent scalar function that controls the noise level along the diffusion process. The induced perturbation kernel has an analytic form as (see the derivation in \cref{app:proof}):
\begin{align}
    p_{0t}({\rvx}^L|{\rvx}^L_0) = \gN({\rvx}^L_t; 
    {\rvx}^L_0 
    e^{-\frac{1}{2}\int_0^t \beta(s) \ud s}, 
    \rmH - \rmH e^{-\int_0^t \beta(s) \ud s}
    ). \nonumber
\end{align}
Given a schedule that satisfies $\lim_{t\to 1} \int_0^t \beta(s) \ud s =\infty$, the above perturbation process arrives at the prior distribution $p_1(\rvx^L_1) \propto \exp(-\frac{1}{2}\rvx_1^{L\intercal}\rmH\rvx^L_1)$ at time $t=1$, which can be efficiently sampled. Intuitively, the anisotropic perturbation process leverages the connection information in the chemical graph $\gG_C$. The bonded atoms are initially set close and then gradually perturbed by correlated noises.

The model is based on an SE(3)-equivariant neural network \cite{satorras2021n,guan2021energy}, incorporating both a k-nearest-neighbor graph built upon the protein-ligand complex and a ligand chemical graph. This design ensures the model is aware of both protein-ligand interactions and atom connections induced by chemical bonds. We leave the details of model architecture design in \cref{app:model_architecture}.

We denote the final output of the SE(3)-equivariant neural network as $\mD_\vtheta(\rvx^L_t,t)$. For simplicity, we use a simple reconstruction loss that is approximately equivariant to the score matching objective as follows:
\begin{align}
    \gL = \mathbb{E}_{t,p_0(\rvx^L_0),p_{0t}(\rvx^L_t|\rvx^L_0)}[\Vert
    \mD_\vtheta(\rvx^L_t,t) - \rvx^L_0
    \Vert^2].
\end{align}
The estimated score function can then be formulated as 
\begin{align}
\textstyle
     \nabla_{\rvx^L} \log p_t(\rvx^L) \approx - \frac{1}{\sqrt{1-\int_0^s \beta (s) \,\ud s}} (\rvx^L - \mD_\vtheta(\rvx^L,t)),
     \nonumber
\end{align}
with which we can generate the ligand poses given the ligand's chemical graph and the receptor's 3D structures by solving the reverse-time SDE as in \cref{eq:reverse_sde}.

\subsection{\seqmodel}
\label{subsec:seqmodel}
We introduce \seqmodel that predicts the ground-truth residue type given the noisy ligands. The model architecture is similar to that of \dockmodel. Differently, the input and output of the model are modified due to the following considerations.

With a structure prediction model in hand, we could still not be able to design cyclic peptides, since their sequence is unknown. This is a classic ``chicken-and-egg" problem. This motivates us to alternately denoise the 3D structures and update the residue types. Since the residue type is determined by the side-chain R group, it will provide a shortcut for the model to predict the ground-truth residue type if the complete chemical graph of the noisy ligands is input. Thus, we remove the side chain of the canonical amino acid residues except for those that are involved in cyclization. 

The model produces a hidden state (i.e., $\rvh$) for each atom. For the $i$-th residue (whose ground-truth amino acid type is $a_i$) in a peptide with $N$ residues, we aggregate the hidden states of the backbone atoms (i.e, $\text{N-C}_{\alpha}\text{-C-O}$) and use a multilayer perceptron (MLP) to predict the amino acid type. The model is trained via the following objective:
\begin{align}
    \textstyle
    \gL = \sum_i^N  - \log p_\mathbf{\phi}(a_i| \mD_\vtheta(\rvx^L_t,t), \gG_C, \gT, t),
\end{align}
where $p_\mathbf{\phi}$ is the model-induced probability for the residue type. \dockmodel (i.e, $\mD_\vtheta$) is first pretrained and then fixed during the training of \seqmodel. 

\subsection{Routed Sampling for Cyclic Peptide Design}
\label{subsec:sampling}

With trained \dockmodel and \seqmodel, we can iteratively update both the sequence and structure through alternate calls to the two models, enabling the generation of all types of cyclic peptides, as shown in \cref{fig:overview}.

Given the number of residues within the peptide and the cyclization information, the atoms in a cyclic peptide can be categorized into two classes: The first class is known in terms of their chemical graph (though their 3D structures are not determined), comprising all backbone atoms and the atoms constrained by cyclization. The second class is unknown, consisting of the side chains of canonical amino acid residues not constrained by cyclization. 
Without loss of generality, consider a specific type of side-to-tail cyclization where the sulfur atom (S) in the side chain of the $i$-th residue and the carboxylic acid (COOH) group of the C-terminus engage in chemical reactions to form a C-S thioester bond. In this scenario, the chemical graphs induced by all backbone atoms (since all canonical amino acid residues share the pattern N-C$_\alpha$-C-O) and all atoms of the cysteine (CYS) providing the sulfur for the C-S bond that forms the cyclic structure are known. Conversely, the chemical graph between other atoms, specifically the side chains of residues except for the aforementioned CYS residue, remains unknown. 
In the remainder of this paper, we refer to the atoms associated with the known chemical graph due to cyclization as \textit{cyclization-constrained} atoms, and the others as \textit{free-residue} atoms.

Inspired by \citet{chu2024all}, for the free-residue part, we maintain an atom73 state (i.e., coordinates), a ``superposition'' for each residue, where all possible amino acid types for this residue share the backbone atoms and C$_\beta$, while possessing unique side chain atoms. For the cyclization-constrained part, we maintain a cyclization-constrained state separately. 
The general idea of routed sampling is to switch to different collapsed (i.e., specific) atom states for the free-residue part and assemble a new valid chemical graph based on the predicted residue type and cyclization information at each step of solving the reverse-time SDE. Specifically, at each step, \dockmodel denoises the atom coordinates, 
 
In practice, for both cyclization-constrained atoms and free-residue atoms, we maintain both denoised states and current states. This approach is needed because, during the sampling process, the cyclization-constrained atoms and backbone atoms are constantly present, while the side-chain atoms of the free residues are sometimes sampled, resulting in them potentially not being adequately updated. Thus, we reuse the previous denoised structure to solve the SDE and align the time (or noise level) of all sampled atoms to the same point. This alignment is necessary because the \dockmodel expects all input atoms to be at the same noise level, especially when a residue type is sampled after not being sampled for a few steps. Please refer to \cref{app:routed_sampling} for more details.

\definecolor{darkgreen}{RGB}{34,139,34}
\begin{table*}[t!]
    \centering
    \caption{Summary of properties of reference peptides, linear peptides designed by baseline methods, and cyclic peptides designed by \method. ($\downarrow$) / ($\uparrow$) denotes a smaller / larger number is better.}
    \renewcommand{\arraystretch}{1.2}
    \begin{tabular}{l|c|l|cc|cc|c}
    \toprule
    \multirow{2}{*}{Method}
    & \multirow{2}{*}{Co-Design}
    & \multirow{2}{*}{Peptide Type} 
    & \multicolumn{2}{c|}{Stability ($\downarrow$)} & \multicolumn{2}{c|}{Affinity ($\downarrow$)} & \multirow{2}{*}{Diversity ($\uparrow$)} \\
    & & & Avg. & Med. & Avg. & Med. &  \\
    \midrule
     {Reference} & N/A &  Linear & -672.53 & -634.71 & -85.03 & -78.70 & N/A \\ 
    \midrule
    RFDiffusion & \XSolidBrush  & Linear  & -633.51 & -607.82 & -70.30 & -61.35 & 0.55 \\
    ProteinGenerator & \Checkmark & Linear & -576.39 & -554.70 & -46.98 & -40.39  & 0.58  \\
    PepFlow & \Checkmark & Linear   &  -576.16 & -498.31 & -47.88 & -42.40 & 0.70 \\
    PepGLAD & \Checkmark & Linear   & -359.44  & -310.33 &  -45.06 &  -38.56 & 0.79 \\
    \midrule
    \method  & \Checkmark & Head-to-tail Cyclic & -568.04 &  -519.66 & -50.86 & -46.62 & 0.79 \\
    \method  & \Checkmark & Head-to-side Cyclic & -564.81 & -508.88 & -49.81 & -44.16 & 0.79 \\
    \method  & \Checkmark & Side-to-tail Cyclic & -547.14 &
    -502.20 & -46.92 & -37.25 & 0.78 \\
    \method  & \Checkmark & Side-to-side Cyclic & -537.09 &  -479.94 & -49.73 & -44.71 & 0.79 \\
    \method  & \Checkmark & Mix & -580.67 & -527.80 & -55.71 & -48.42 & 0.79 \\
    
    \bottomrule
    \end{tabular}\label{tab:main_table}
    \renewcommand{\arraystretch}{1}
\end{table*}

Notably, the two models, \dockmodel and \seqmodel, do not account for the atom partition introduced by amino acid residues, as they model atoms and bonds at the most fundamental level. The concept of amino acid residues is introduced only during routed sampling to ensure that the free-residue part remains a canonical residue rather than an arbitrary molecule.

\section{Experiments}
\label{sec:exp}

\begin{figure*}[t!]
\begin{center}
\centerline{\includegraphics[width=0.94\textwidth]{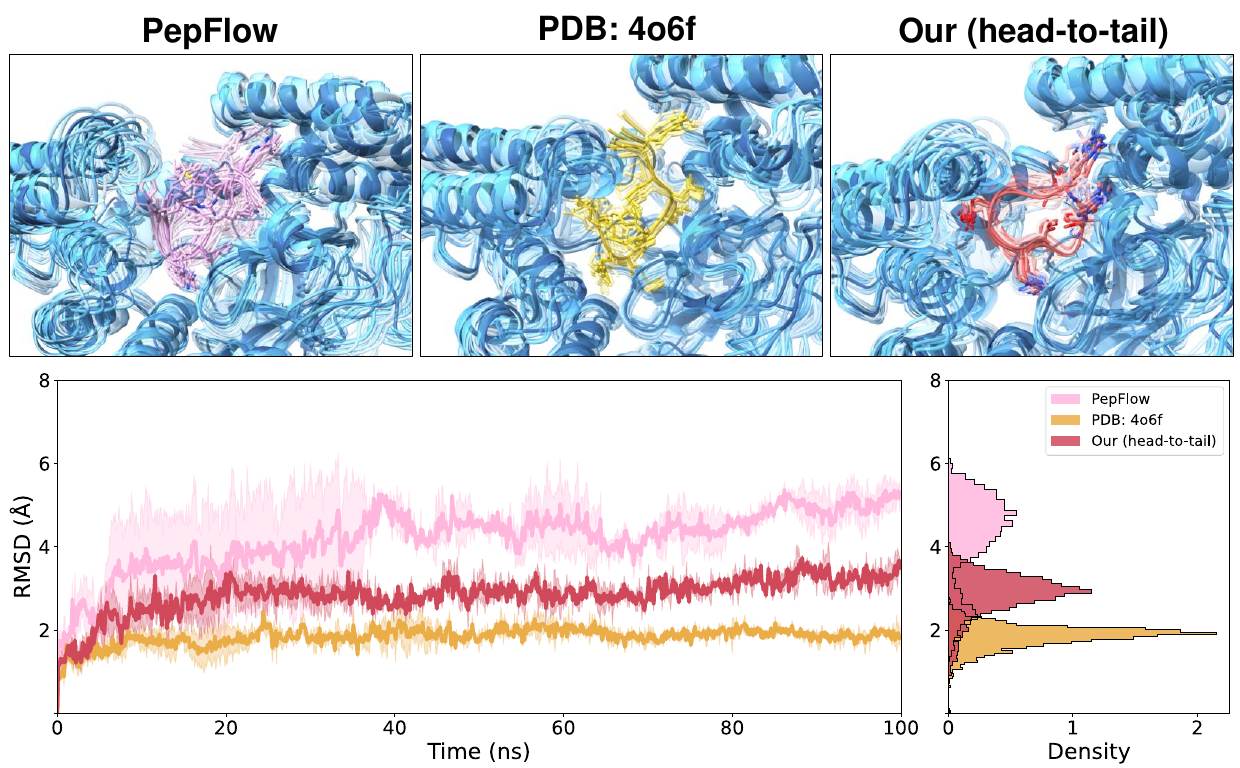}}
\vspace{-4mm}
\caption{Discovery  of  new  SMYD2 cyclic peptide inhibitors via applications of \method. Upper: Visualization of conformational ensembles for the designed peptides sampled by MD. Bottom: RMSD analysis of all heavy atoms within the designed peptides.
}
\label{fig:md_case_4o6f}
\vspace{-6mm}
\end{center}
\end{figure*}

\subsection{Experimental Setup}
\label{subsec:experimental_setup}
\textbf{Dataset.} We have curated two datasets of protein-ligand complexes featuring small molecules and peptides as ligands, respectively. All complexes with atoms whose elements are beyond \{C, N, O, F, S, Cl, Se, Br\} are not included.
The small molecule dataset is sourced from PDBBind \citep{wang2005pdbbind} and has 14,348 protein-ligand complexes.
The peptide dataset is derived from RCSB PDB~\citep{burley2023rcsb}, Propedia~\citep{martins2023propedia} and PepBDB~\citep{wen2019pepbdb}. It comprises 20,033 protein-ligand complexes, featuring peptide ligands composed of fewer than 30 residues.
Samples are clustered by receptor sequence identity of $0.3$ to split the dataset into training and validation sets.
To train \dockmodel, we utilize the curated small molecule dataset and a subset of the peptide dataset containing ligands with fewer than 200 heavy atoms.
To train \seqmodel, we use the whole peptide dataset.
Please refer to \cref{app:peptide_dataset} for more details about data.

\textbf{Baselines.} As our method is the first cyclic peptide design method based on generative models, we compare our approach with various established methods for linear peptide design:  \textbf{RFDiffusion} \citep{watson2023novo} generates protein backbones, and sequences are later predicted by ProteinMPNN~\citep{dauparas2022robust}; \textbf{ProteinGenerator} \citep{lisanza2024multistate} improves RFDiffusion by jointly sampling backbones and corresponding sequences; \textbf{PepFlow} \citep{li2024full} is a flow-based full-atom peptide generative model that generates the translation, rotation, and side-chain torsion angles of each residue frame within a peptide; \textbf{PepGLAD} \citep{kong2024fullatom} is a full-atom peptide design method that utilizes a latent diffusion model.

\textbf{Evaluation.} We use Rosetta~\citep{chaudhury2010pyrosetta} to compute the total energy of reference ligands, linear peptides designed by baseline methods, and cyclic peptides engineered by our approaches. This energy measurement serves as an indicator of the stability (or rationality) of a peptide's 3D binding pose. Hence, we define this energy metric as \textbf{Stability}. We also evaluate the interface binding energy, a crucial metric that indicates the binding affinity of the ligand peptide to its receptor. This assessment is essential for evaluating the ligand peptide's functionality, particularly when designing peptides for therapeutic applications. We denote this type of energy as \textbf{Affinity}.
We also report \textbf{Diversity}, the average of one minus the pair-wise TM-Score \citep{zhang2005tm} among the designed peptides, reflecting structural dissimilarities.
As the reference ligands for the targets in the test set are linear, we do not report metrics that necessitate a reference sequence or structure, such as Amino Acid Recovery (AAR) and Root Mean Square Deviation (RMSD). 
We selected 100 protein pockets with a large volume for testing. Specifically, these targets have receptors with more than 1,000 surrounding atoms around the reference peptide ligands, serving as reliable indicators of adequate volume.
For each target, all peptide design methods are used to generate a batch of peptides, from which we select the most promising (i.e., lowest energy) peptide ligand. 
We design four types of cyclic types (head-to-tail, head-to-side, side-to-side, and side-to-tail) using our methods, respectively, and we also report the mixed results, where the best promising peptide ligands might exhibit different cyclization types.
We then report the average and median metrics across all targets. Please refer to \cref{app:experimental_details} for more details. This evaluation strategy mirrors practical drug design scenarios where the leading ligand candidates are identified for advancement to the subsequent stages of drug development.  

\subsection{Main Results}

The results are shown in \cref{tab:main_table}. Among all co-design methods, our method exhibits superior energy performance in terms of both stability and affinity and also best diversity. Interestingly, we find that head-to-tail and head-to-side cyclic peptides show better performance than side-to-tail and side-to-side cyclic peptides. This might be due to the training dataset where C-N bonds (main covalent bonds that form head-to-tail and head-to-side cyclic structures) are more frequent than S-S and C-S bonds (main covalent bonds that form side-to-tail and side-to-side cyclic structures). This aligns with the fact that C-N bonds are generally more stable than S-S and C-S bonds in the physical world. Among all methods, RFDiffusion shows the best energy performance but low diversity, as it tends to generate $\alpha$-helices in a certain pattern. See \cref{app:ablation_studies}
for ablation studies.

\subsection{Case Studies}

\begin{figure*}[t!]
\begin{center}
\centerline{\includegraphics[width=0.94\textwidth]{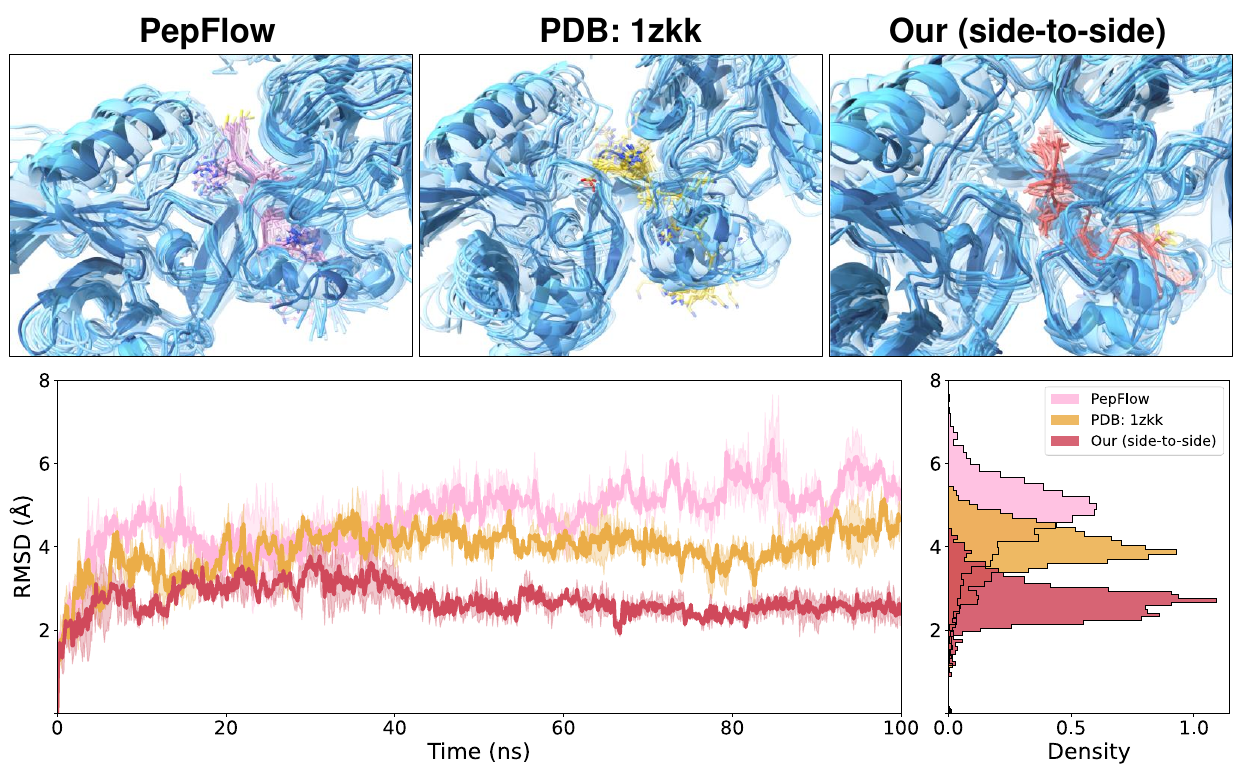}}
\vspace{-4mm}
\caption{Discovery  of  new  SET8 cyclic peptide inhibitors via applications of \method.
Upper: Visualization of conformational ensembles for the designed peptides sampled by MD. Bottom: RMSD analysis of all heavy atoms within the designed peptides.
}
\label{fig:md_case_1zkk}
\vspace{-6mm}
\end{center}
\end{figure*}

Here, we demonstrate how \method can be seamlessly integrated into real-world cyclic peptide design pipelines and discuss two scenarios: design of SMYD2 peptide inhibitors via head-to-tail cyclization and SET8 inhibitors via side-to-side cyclization. 

\textbf{Design of SMYD2 peptide inhibitors via head-to-tail cyclization.} SMYD2 is an oncogene that critically regulates tumor-related signaling pathways, making it an attractive target for cancer therapy~\citep{zheng2022protein}. In particular, SMYD2 attenuates estrogen signaling by weakening ER$\alpha$-dependent transactivation~\citep{zhang2013regulation}. Insight into this interaction comes from the SMYD2–ER$\alpha$ co-crystal structure (PDB: 4O6F), which reveals ER$\alpha$ in a U-shaped conformation nestled within SMYD2’s deep binding pocket~\citep{jiang2014structural}. The distance between ER$\alpha$’s N-terminal (head) and C-terminal (tail) in this complex is only 5.9 \AA, a finding that has inspired the design of rigid head-to-tail cyclic peptides to capitalize on this proximity and enhance binding affinity.

Herein, \method was employed to design new SMYD2 inhibitors by generating head-to-tail cyclic peptides. The binding site of ER$\alpha$ (PDB: 4o6f) was selected as the active pocket, and 8 cyclic peptides were generated through head-to-tail cyclization. All peptides demonstrated favorable Rosetta affinity scores (\cref{fig:4o6f_select}), with H2T-6 achieving the most favorable score of -33.9 kcal/mol. To further accurately assess its binding stability and affinity, 100 ns molecular dynamics (MD) simulations were performed on the H2T-6-SMYD2 complex. More details on the system preparation and simulation protocol of MD are in Appendix~\ref{app:md_setup}. The simulations also included the crystallized linear peptide (ground truth) and the PepFlow-generated linear peptide for comparison. Each simulation was repeated twice to ensure consistency. As shown in \cref{fig:md_case_4o6f}, the linear peptide generated by PepFlow exhibited higher flexibility, with an average peptide RMSD of 4.59 \AA\ over the last 50 ns of equilibrated trajectories. In contrast, H2T-6 displayed an average peptide RMSD of 3.05 \AA, comparable to the ground truth linear peptide (1.92 \AA). More importantly, binding free energy analysis using MM-PBSA~\citep{wang2019end} revealed that H2T-6 had the highest binding affinity (-24.02 kcal/mol), outperforming both the ground truth peptide (-19.00 kcal/mol) and the PepFlow-generated peptide (-7.26 kcal/mol). These results suggest that H2T-6 maintains a stable binding conformation which also exhibits high binding affinity, indicating its potential as a candidate SMYD2 inhibitor for further investigation.

\textbf{Design of SET8 peptide inhibitors via side-to-side cyclization.} SET8 is the only lysine methyltransferase that specifically catalyzes the methylation of histone H4 at the 20th lysine~\citep{qian2006set}. SET8-mediated protein modifications are involved in numerous physiological processes, and its dysregulation is closely linked to various human diseases, particularly cancer development and prognosis~\citep{yang2021histone}. 

Again, we applied \method to design seed cyclic peptide inhibitors targeting SET8. The 3D structure of SET8 in complex with an H4 peptide (PDB: 1zkk) was used, with the H4 binding site selected as the active pocket~\citep{couture2005structural}. To explore an alternative cyclization approach, we employed a widely used side-to-side strategy to generate 8 cyclic peptides. Likely, we picked up the top cyclic peptide, S2S-4 (see \cref{fig:1zkk_select}), based on its Rosetta affinity score, and performed 100 ns MD simulations with 2 repeats, along with two reference linear peptides. The RMSD and MM-PBSA analyses revealed that S2S-4 not only demonstrated a significantly lower peptide RMSD (2.54 \AA) compared to the reference linear peptides (ground truth: 4.06 \AA; PepFlow: 5.23 \AA), but also exhibited a lower binding free energy (S2S-4: -12.48 kcal/mol; ground truth: -6.39 kcal/mol; PepFlow: -9.26 kcal/mol). These results indicate that S2S-4 might serve as a potential candidate for later studies on SET8 inhibition.

\section{Conclusion}
In conclusion, \method is a generative algorithm capable of producing diverse types of cyclic peptides given 3D receptor structures, thereby paving the way for advancements in peptide-based drug discovery. Our approach enhances drug development by designing stable and high-affinity cyclic peptides. Case studies supported by molecular dynamics simulations validate its practical utility and real-world applicability. Limitations and future work are discussed in \cref{app:limitations}.

\section*{Acknowledgments}
We thank anonymous reviewers for their insightful feedback. 
We would like to extend our gratitude to Yi Zhou and Lihao Wang for their valuable feedback on our methodology, as well as to Ellen Wang, Tianze Zheng and Wen Yan for their assistance with the molecular dynamics simulations.

\section*{Impact Statement}

\method contributes to the advancement of computational biology, particularly in the field of cyclic peptide design, by addressing key challenges that have hindered progress in this area. By integrating generative structure prediction and sequence prediction, our approach enables the design of stable and high-affinity cyclic peptides, offering valuable applications in drug discovery and therapeutic development. While our primary focus is on the positive applications of \method, such as developing novel peptide-based treatments, we acknowledge the ethical considerations associated with any powerful generative tool. There is a potential risk of misuse, including the design of harmful bioactive compounds. To mitigate such concerns, we emphasize the responsible use of our method and encourage the scientific community to apply it for constructive and beneficial purposes. Our commitment to ethical research practices ensures that \method serves as a force for innovation in pharmaceutical development while upholding societal safety and integrity.

\clearpage
\bibliography{main}

\begin{thebibliography}{120}
\providecommand{\natexlab}[1]{#1}
\providecommand{\url}[1]{\texttt{#1}}
\expandafter\ifx\csname urlstyle\endcsname\relax
  \providecommand{\doi}[1]{doi: #1}\else
  \providecommand{\doi}{doi: \begingroup \urlstyle{rm}\Url}\fi

\bibitem[Abramson et~al.(2024)Abramson, Adler, Dunger, Evans, Green, Pritzel, Ronneberger, Willmore, Ballard, Bambrick, et~al.]{abramson2024accurate}
Abramson, J., Adler, J., Dunger, J., Evans, R., Green, T., Pritzel, A., Ronneberger, O., Willmore, L., Ballard, A.~J., Bambrick, J., et~al.
\newblock Accurate structure prediction of biomolecular interactions with alphafold 3.
\newblock \emph{Nature}, pp.\  1--3, 2024.

\bibitem[Alford et~al.(2017)Alford, Leaver-Fay, Jeliazkov, O’Meara, DiMaio, Park, Shapovalov, Renfrew, Mulligan, Kappel, et~al.]{alford2017rosetta}
Alford, R.~F., Leaver-Fay, A., Jeliazkov, J.~R., O’Meara, M.~J., DiMaio, F.~P., Park, H., Shapovalov, M.~V., Renfrew, P.~D., Mulligan, V.~K., Kappel, K., et~al.
\newblock The rosetta all-atom energy function for macromolecular modeling and design.
\newblock \emph{Journal of chemical theory and computation}, 13\penalty0 (6):\penalty0 3031--3048, 2017.

\bibitem[Baek et~al.(2021)Baek, DiMaio, Anishchenko, Dauparas, Ovchinnikov, Lee, Wang, Cong, Kinch, Schaeffer, et~al.]{baek2021accurate}
Baek, M., DiMaio, F., Anishchenko, I., Dauparas, J., Ovchinnikov, S., Lee, G.~R., Wang, J., Cong, Q., Kinch, L.~N., Schaeffer, R.~D., et~al.
\newblock Accurate prediction of protein structures and interactions using a three-track neural network.
\newblock \emph{Science}, 373\penalty0 (6557):\penalty0 871--876, 2021.

\bibitem[Berman et~al.(2000)Berman, Westbrook, Feng, Gilliland, Bhat, Weissig, Shindyalov, and Bourne]{berman2000protein}
Berman, H.~M., Westbrook, J., Feng, Z., Gilliland, G., Bhat, T.~N., Weissig, H., Shindyalov, I.~N., and Bourne, P.~E.
\newblock The protein data bank.
\newblock \emph{Nucleic acids research}, 28\penalty0 (1):\penalty0 235--242, 2000.

\bibitem[Bhardwaj et~al.(2016)Bhardwaj, Mulligan, Bahl, Gilmore, Harvey, Cheneval, Buchko, Pulavarti, Kaas, Eletsky, et~al.]{bhardwaj2016accurate}
Bhardwaj, G., Mulligan, V.~K., Bahl, C.~D., Gilmore, J.~M., Harvey, P.~J., Cheneval, O., Buchko, G.~W., Pulavarti, S.~V., Kaas, Q., Eletsky, A., et~al.
\newblock Accurate de novo design of hyperstable constrained peptides.
\newblock \emph{Nature}, 538\penalty0 (7625):\penalty0 329--335, 2016.

\bibitem[Bhardwaj et~al.(2022)Bhardwaj, O’Connor, Rettie, Huang, Ramelot, Mulligan, Alpkilic, Palmer, Bera, Bick, et~al.]{bhardwaj2022accurate}
Bhardwaj, G., O’Connor, J., Rettie, S., Huang, Y.-H., Ramelot, T.~A., Mulligan, V.~K., Alpkilic, G.~G., Palmer, J., Bera, A.~K., Bick, M.~J., et~al.
\newblock Accurate de novo design of membrane-traversing macrocycles.
\newblock \emph{Cell}, 185\penalty0 (19):\penalty0 3520--3532, 2022.

\bibitem[Boom et~al.(2024)Boom, Greenig, Sormanni, and Liò]{boom2024scorebasedgenerativemodelsdesigning}
Boom, J.~D., Greenig, M., Sormanni, P., and Liò, P.
\newblock Score-based generative models for designing binding peptide backbones, 2024.
\newblock URL \url{https://arxiv.org/abs/2310.07051}.

\bibitem[Buckton et~al.(2021)Buckton, Rahimi, and McAlpine]{buckton2021cyclicas}
Buckton, L.~K., Rahimi, M.~N., and McAlpine, S.~R.
\newblock Cyclic peptides as drugs for intracellular targets: the next frontier in peptide therapeutic development.
\newblock \emph{Chemistry--A European Journal}, 27\penalty0 (5):\penalty0 1487--1513, 2021.

\bibitem[Burley et~al.(2023)Burley, Bhikadiya, Bi, Bittrich, Chao, Chen, Craig, Crichlow, Dalenberg, Duarte, et~al.]{burley2023rcsb}
Burley, S.~K., Bhikadiya, C., Bi, C., Bittrich, S., Chao, H., Chen, L., Craig, P.~A., Crichlow, G.~V., Dalenberg, K., Duarte, J.~M., et~al.
\newblock Rcsb protein data bank (rcsb. org): delivery of experimentally-determined pdb structures alongside one million computed structure models of proteins from artificial intelligence/machine learning.
\newblock \emph{Nucleic acids research}, 51\penalty0 (D1):\penalty0 D488--D508, 2023.

\bibitem[Camarero \& Muir(1999)Camarero and Muir]{camarero1999biosynthesis}
Camarero, J.~A. and Muir, T.~W.
\newblock Biosynthesis of a head-to-tail cyclized protein with improved biological activity.
\newblock \emph{Journal of the American Chemical Society}, 121\penalty0 (23):\penalty0 5597--5598, 1999.

\bibitem[Campbell et~al.(2024)Campbell, Yim, Barzilay, Rainforth, and Jaakkola]{campbell2024generative}
Campbell, A., Yim, J., Barzilay, R., Rainforth, T., and Jaakkola, T.
\newblock Generative flows on discrete state-spaces: Enabling multimodal flows with applications to protein co-design.
\newblock In \emph{Forty-first International Conference on Machine Learning}, 2024.

\bibitem[Cao et~al.(2024)Cao, Chen, Zhang, Wang, Huang, Yu, Jiang, Fan, Zhang, Zhou, et~al.]{cao2024surfdock}
Cao, D., Chen, M., Zhang, R., Wang, Z., Huang, M., Yu, J., Jiang, X., Fan, Z., Zhang, W., Zhou, H., et~al.
\newblock Surfdock is a surface-informed diffusion generative model for reliable and accurate protein--ligand complex prediction.
\newblock \emph{Nature Methods}, pp.\  1--13, 2024.

\bibitem[Cao et~al.(2022)Cao, Coventry, Goreshnik, Huang, Sheffler, Park, Jude, Markovi{\'c}, Kadam, Verschueren, et~al.]{cao2022design}
Cao, L., Coventry, B., Goreshnik, I., Huang, B., Sheffler, W., Park, J.~S., Jude, K.~M., Markovi{\'c}, I., Kadam, R.~U., Verschueren, K.~H., et~al.
\newblock Design of protein-binding proteins from the target structure alone.
\newblock \emph{Nature}, 605\penalty0 (7910):\penalty0 551--560, 2022.

\bibitem[Caplin et~al.(2014)Caplin, Pavel, {\'C}wik{\l}a, Phan, Raderer, Sedl{\'a}{\v{c}}kov{\'a}, Cadiot, Wolin, Capdevila, Wall, et~al.]{caplin2014lanreotide}
Caplin, M.~E., Pavel, M., {\'C}wik{\l}a, J.~B., Phan, A.~T., Raderer, M., Sedl{\'a}{\v{c}}kov{\'a}, E., Cadiot, G., Wolin, E.~M., Capdevila, J., Wall, L., et~al.
\newblock Lanreotide in metastatic enteropancreatic neuroendocrine tumors.
\newblock \emph{New England Journal of Medicine}, 371\penalty0 (3):\penalty0 224--233, 2014.

\bibitem[Chaudhury et~al.(2010)Chaudhury, Lyskov, and Gray]{chaudhury2010pyrosetta}
Chaudhury, S., Lyskov, S., and Gray, J.~J.
\newblock Pyrosetta: a script-based interface for implementing molecular modeling algorithms using rosetta.
\newblock \emph{Bioinformatics}, 26\penalty0 (5):\penalty0 689--691, 2010.

\bibitem[Chen et~al.(2025)Chen, Xue, Zhou, Zheng, Zeng, and Gu]{chen2025apm}
Chen, R., Xue, D., Zhou, X., Zheng, Z., Zeng, X., and Gu, Q.
\newblock An all-atom generative model for designing protein complexes.
\newblock In \emph{International Conference on Machine Learning}, 2025.

\bibitem[Chen et~al.(2024)Chen, Pertsemlidis, and Chatterjee]{chen2024pepmlm}
Chen, T., Pertsemlidis, S., and Chatterjee, P.
\newblock Pep{MLM}: Target sequence-conditioned generation of peptide binders via masked language modeling.
\newblock In \emph{ICLR 2024 Workshop on Generative and Experimental Perspectives for Biomolecular Design}, 2024.
\newblock URL \url{https://openreview.net/forum?id=p6fz0rq7zu}.

\bibitem[Cheng et~al.(2024)Cheng, Zhou, Yang, Bao, and Gu]{cheng2024decomposed}
Cheng, X., Zhou, X., Yang, Y., Bao, Y., and Gu, Q.
\newblock Decomposed direct preference optimization for structure-based drug design.
\newblock \emph{arXiv preprint arXiv:2407.13981}, 2024.

\bibitem[Chu et~al.(2024)Chu, Kim, Cheng, El~Nesr, Xu, Shuai, and Huang]{chu2024all}
Chu, A.~E., Kim, J., Cheng, L., El~Nesr, G., Xu, M., Shuai, R.~W., and Huang, P.-S.
\newblock An all-atom protein generative model.
\newblock \emph{Proceedings of the National Academy of Sciences}, 121\penalty0 (27):\penalty0 e2311500121, 2024.

\bibitem[Corso et~al.(2023)Corso, St{\"a}rk, Jing, Barzilay, and Jaakkola]{corsodiffdock}
Corso, G., St{\"a}rk, H., Jing, B., Barzilay, R., and Jaakkola, T.~S.
\newblock Diffdock: Diffusion steps, twists, and turns for molecular docking.
\newblock In \emph{The Eleventh International Conference on Learning Representations}, 2023.

\bibitem[Costa et~al.(2023)Costa, Sousa, and Fernandes]{costa2023cyclic}
Costa, L., Sousa, E., and Fernandes, C.
\newblock Cyclic peptides in pipeline: what future for these great molecules?
\newblock \emph{Pharmaceuticals}, 16\penalty0 (7):\penalty0 996, 2023.

\bibitem[Couture et~al.(2005)Couture, Collazo, Brunzelle, and Trievel]{couture2005structural}
Couture, J.-F., Collazo, E., Brunzelle, J.~S., and Trievel, R.~C.
\newblock Structural and functional analysis of set8, a histone h4 lys-20 methyltransferase.
\newblock \emph{Genes \& development}, 19\penalty0 (12):\penalty0 1455--1465, 2005.

\bibitem[Dauparas et~al.(2022)Dauparas, Anishchenko, Bennett, Bai, Ragotte, Milles, Wicky, Courbet, de~Haas, Bethel, et~al.]{dauparas2022robust}
Dauparas, J., Anishchenko, I., Bennett, N., Bai, H., Ragotte, R.~J., Milles, L.~F., Wicky, B.~I., Courbet, A., de~Haas, R.~J., Bethel, N., et~al.
\newblock Robust deep learning--based protein sequence design using proteinmpnn.
\newblock \emph{Science}, 378\penalty0 (6615):\penalty0 49--56, 2022.

\bibitem[Dhariwal \& Nichol(2021)Dhariwal and Nichol]{dhariwal2021diffusion}
Dhariwal, P. and Nichol, A.
\newblock Diffusion models beat gans on image synthesis.
\newblock \emph{Advances in neural information processing systems}, 34:\penalty0 8780--8794, 2021.

\bibitem[Dolinsky et~al.(2007)Dolinsky, Czodrowski, Li, Nielsen, Jensen, Klebe, and Baker]{dolinsky2007pdb2pqr}
Dolinsky, T.~J., Czodrowski, P., Li, H., Nielsen, J.~E., Jensen, J.~H., Klebe, G., and Baker, N.~A.
\newblock Pdb2pqr: expanding and upgrading automated preparation of biomolecular structures for molecular simulations.
\newblock \emph{Nucleic acids research}, 35\penalty0 (suppl\_2):\penalty0 W522--W525, 2007.

\bibitem[Driggers et~al.(2008)Driggers, Hale, Lee, and Terrett]{driggers2008exploration}
Driggers, E.~M., Hale, S.~P., Lee, J., and Terrett, N.~K.
\newblock The exploration of macrocycles for drug discovery—an underexploited structural class.
\newblock \emph{Nature Reviews Drug Discovery}, 7\penalty0 (7):\penalty0 608--624, 2008.

\bibitem[Eberhardt et~al.(2021)Eberhardt, Santos-Martins, Tillack, and Forli]{eberhardt2021autodock}
Eberhardt, J., Santos-Martins, D., Tillack, A.~F., and Forli, S.
\newblock Autodock vina 1.2. 0: New docking methods, expanded force field, and python bindings.
\newblock \emph{Journal of chemical information and modeling}, 61\penalty0 (8):\penalty0 3891--3898, 2021.

\bibitem[Fang et~al.(2024)Fang, Pang, Xuan, Chan, and Leung]{fang2024recent}
Fang, P., Pang, W.-K., Xuan, S., Chan, W.-L., and Leung, K. C.-F.
\newblock Recent advances in peptide macrocyclization strategies.
\newblock \emph{Chemical Society Reviews}, 2024.

\bibitem[Fass(2012)]{fass2012disulfide}
Fass, D.
\newblock Disulfide bonding in protein biophysics.
\newblock \emph{Annual review of biophysics}, 41\penalty0 (1):\penalty0 63--79, 2012.

\bibitem[Friesner et~al.(2004)Friesner, Banks, Murphy, Halgren, Klicic, Mainz, Repasky, Knoll, Shelley, Perry, et~al.]{friesner2004glide}
Friesner, R.~A., Banks, J.~L., Murphy, R.~B., Halgren, T.~A., Klicic, J.~J., Mainz, D.~T., Repasky, M.~P., Knoll, E.~H., Shelley, M., Perry, J.~K., et~al.
\newblock Glide: a new approach for rapid, accurate docking and scoring. 1. method and assessment of docking accuracy.
\newblock \emph{Journal of medicinal chemistry}, 47\penalty0 (7):\penalty0 1739--1749, 2004.

\bibitem[Gao et~al.(2023{\natexlab{a}})Gao, Tan, Chen, Zhang, Xia, Li, and Li]{gao2023kw}
Gao, Z., Tan, C., Chen, X., Zhang, Y., Xia, J., Li, S., and Li, S.~Z.
\newblock Kw-design: Pushing the limit of protein design via knowledge refinement.
\newblock In \emph{The Twelfth International Conference on Learning Representations}, 2023{\natexlab{a}}.

\bibitem[Gao et~al.(2023{\natexlab{b}})Gao, Tan, and Li]{gaopifold}
Gao, Z., Tan, C., and Li, S.~Z.
\newblock Pifold: Toward effective and efficient protein inverse folding.
\newblock In \emph{The Eleventh International Conference on Learning Representations}, 2023{\natexlab{b}}.

\bibitem[Garcia~Jimenez et~al.(2023)Garcia~Jimenez, Poongavanam, and Kihlberg]{garcia2023macrocycles}
Garcia~Jimenez, D., Poongavanam, V., and Kihlberg, J.
\newblock Macrocycles in drug discovery-learning from the past for the future.
\newblock \emph{Journal of Medicinal Chemistry}, 66\penalty0 (8):\penalty0 5377--5396, 2023.

\bibitem[Giordanetto \& Kihlberg(2014)Giordanetto and Kihlberg]{giordanetto2014macrocyclic}
Giordanetto, F. and Kihlberg, J.
\newblock Macrocyclic drugs and clinical candidates: what can medicinal chemists learn from their properties?
\newblock \emph{Journal of medicinal chemistry}, 57\penalty0 (2):\penalty0 278--295, 2014.

\bibitem[Guan et~al.(2021)Guan, Qian, Ma, Ma, and Peng]{guan2021energy}
Guan, J., Qian, W.~W., Ma, W.-Y., Ma, J., and Peng, J.
\newblock Energy-inspired molecular conformation optimization.
\newblock In \emph{international conference on learning representations}, 2021.

\bibitem[Guan et~al.(2025)Guan, Li, Zhou, Peng, Wang, Luo, Peng, and Ma]{guan2025group}
Guan, J., Li, J., Zhou, X., Peng, X., Wang, S., Luo, Y., Peng, J., and Ma, J.
\newblock Group ligands docking to protein pockets.
\newblock \emph{arXiv preprint arXiv:2501.15055}, 2025.

\bibitem[Ho \& Salimans(2022)Ho and Salimans]{ho2022classifier}
Ho, J. and Salimans, T.
\newblock Classifier-free diffusion guidance.
\newblock \emph{arXiv preprint arXiv:2207.12598}, 2022.

\bibitem[Ho et~al.(2020)Ho, Jain, and Abbeel]{ho2020denoising}
Ho, J., Jain, A., and Abbeel, P.
\newblock Denoising diffusion probabilistic models.
\newblock \emph{Advances in neural information processing systems}, 33:\penalty0 6840--6851, 2020.

\bibitem[Hopkins et~al.(2015)Hopkins, Le~Grand, Walker, and Roitberg]{hopkins2015long}
Hopkins, C.~W., Le~Grand, S., Walker, R.~C., and Roitberg, A.~E.
\newblock Long-time-step molecular dynamics through hydrogen mass repartitioning.
\newblock \emph{Journal of chemical theory and computation}, 11\penalty0 (4):\penalty0 1864--1874, 2015.

\bibitem[Hosseinzadeh et~al.(2021)Hosseinzadeh, Watson, Craven, Li, Rettie, Pardo-Avila, Bera, Mulligan, Lu, Ford, et~al.]{hosseinzadeh2021anchor}
Hosseinzadeh, P., Watson, P.~R., Craven, T.~W., Li, X., Rettie, S., Pardo-Avila, F., Bera, A.~K., Mulligan, V.~K., Lu, P., Ford, A.~S., et~al.
\newblock Anchor extension: a structure-guided approach to design cyclic peptides targeting enzyme active sites.
\newblock \emph{Nature Communications}, 12\penalty0 (1):\penalty0 3384, 2021.

\bibitem[Hsu et~al.(2022)Hsu, Verkuil, Liu, Lin, Hie, Sercu, Lerer, and Rives]{hsu2022learning}
Hsu, C., Verkuil, R., Liu, J., Lin, Z., Hie, B., Sercu, T., Lerer, A., and Rives, A.
\newblock Learning inverse folding from millions of predicted structures.
\newblock In \emph{International conference on machine learning}, pp.\  8946--8970. PMLR, 2022.

\bibitem[Huang et~al.(2024)Huang, Zhang, Wu, Tan, Lin, Gao, Li, and Li]{huang2024redock}
Huang, Y., Zhang, O., Wu, L., Tan, C., Lin, H., Gao, Z., Li, S., and Li, S.~Z.
\newblock Re-dock: Towards flexible and realistic molecular docking with diffusion bridge.
\newblock In \emph{Forty-first International Conference on Machine Learning}, 2024.
\newblock URL \url{https://openreview.net/forum?id=QRjTDhCIO8}.

\bibitem[Ingraham et~al.(2019)Ingraham, Garg, Barzilay, and Jaakkola]{ingraham2019generative}
Ingraham, J., Garg, V., Barzilay, R., and Jaakkola, T.
\newblock Generative models for graph-based protein design.
\newblock \emph{Advances in neural information processing systems}, 32, 2019.

\bibitem[Jiang et~al.(2014)Jiang, Trescott, Holcomb, Zhang, Brunzelle, Sirinupong, Shi, and Yang]{jiang2014structural}
Jiang, Y., Trescott, L., Holcomb, J., Zhang, X., Brunzelle, J., Sirinupong, N., Shi, X., and Yang, Z.
\newblock Structural insights into estrogen receptor $\alpha$ methylation by histone methyltransferase smyd2, a cellular event implicated in estrogen signaling regulation.
\newblock \emph{Journal of molecular biology}, 426\penalty0 (20):\penalty0 3413--3425, 2014.

\bibitem[Jin et~al.(2022)Jin, Wohlwend, Barzilay, and Jaakkola]{jin2022iterative}
Jin, W., Wohlwend, J., Barzilay, R., and Jaakkola, T.~S.
\newblock Iterative refinement graph neural network for antibody sequence-structure co-design.
\newblock In \emph{International Conference on Learning Representations}, 2022.
\newblock URL \url{https://openreview.net/forum?id=LI2bhrE_2A}.

\bibitem[Jing et~al.(2021)Jing, Eismann, Suriana, Townshend, and Dror]{jing2021learning}
Jing, B., Eismann, S., Suriana, P., Townshend, R. J.~L., and Dror, R.
\newblock Learning from protein structure with geometric vector perceptrons.
\newblock In \emph{International Conference on Learning Representations}, 2021.
\newblock URL \url{https://openreview.net/forum?id=1YLJDvSx6J4}.

\bibitem[Jing et~al.(2023)Jing, Erives, Pao-Huang, Corso, Berger, and Jaakkola]{jing2023eigenfold}
Jing, B., Erives, E., Pao-Huang, P., Corso, G., Berger, B., and Jaakkola, T.~S.
\newblock Eigenfold: Generative protein structure prediction with diffusion models.
\newblock In \emph{ICLR 2023-Machine Learning for Drug Discovery workshop}, 2023.

\bibitem[Jorgensen et~al.(1983)Jorgensen, Chandrasekhar, Madura, Impey, and Klein]{jorgensen1983comparison}
Jorgensen, W.~L., Chandrasekhar, J., Madura, J.~D., Impey, R.~W., and Klein, M.~L.
\newblock Comparison of simple potential functions for simulating liquid water.
\newblock \emph{The Journal of chemical physics}, 79\penalty0 (2):\penalty0 926--935, 1983.

\bibitem[Kale et~al.(2018)Kale, Villequey, Kong, Zorzi, Deyle, and Heinis]{kale2018cyclization}
Kale, S.~S., Villequey, C., Kong, X.-D., Zorzi, A., Deyle, K., and Heinis, C.
\newblock Cyclization of peptides with two chemical bridges affords large scaffold diversities.
\newblock \emph{Nature chemistry}, 10\penalty0 (7):\penalty0 715--723, 2018.

\bibitem[Kawamura et~al.(2017)Kawamura, M{\"u}nzel, Kojima, Yapp, Bhushan, Goto, Tumber, Katoh, King, Passioura, et~al.]{kawamura2017highly}
Kawamura, A., M{\"u}nzel, M., Kojima, T., Yapp, C., Bhushan, B., Goto, Y., Tumber, A., Katoh, T., King, O.~N., Passioura, T., et~al.
\newblock Highly selective inhibition of histone demethylases by de novo macrocyclic peptides.
\newblock \emph{Nature communications}, 8\penalty0 (1):\penalty0 14773, 2017.

\bibitem[Kong et~al.(2023)Kong, Huang, and Liu]{kong2023end}
Kong, X., Huang, W., and Liu, Y.
\newblock End-to-end full-atom antibody design.
\newblock In \emph{Proceedings of the 40th International Conference on Machine Learning}, pp.\  17409--17429, 2023.

\bibitem[Kong et~al.(2024)Kong, Jia, Huang, and Liu]{kong2024fullatom}
Kong, X., Jia, Y., Huang, W., and Liu, Y.
\newblock Full-atom peptide design with geometric latent diffusion.
\newblock In \emph{The Thirty-eighth Annual Conference on Neural Information Processing Systems}, 2024.
\newblock URL \url{https://openreview.net/forum?id=IAQNJUJe8q}.

\bibitem[Krishna et~al.(2024)Krishna, Wang, Ahern, Sturmfels, Venkatesh, Kalvet, Lee, Morey-Burrows, Anishchenko, Humphreys, et~al.]{krishna2024generalized}
Krishna, R., Wang, J., Ahern, W., Sturmfels, P., Venkatesh, P., Kalvet, I., Lee, G.~R., Morey-Burrows, F.~S., Anishchenko, I., Humphreys, I.~R., et~al.
\newblock Generalized biomolecular modeling and design with rosettafold all-atom.
\newblock \emph{Science}, 384\penalty0 (6693):\penalty0 eadl2528, 2024.

\bibitem[Kulyt{\.e} et~al.(2024)Kulyt{\.e}, Vargas, Mathis, Wang, Hern{\'a}ndez-Lobato, and Li{\`o}]{kulyte2024improving}
Kulyt{\.e}, P., Vargas, F., Mathis, S.~V., Wang, Y.~G., Hern{\'a}ndez-Lobato, J.~M., and Li{\`o}, P.
\newblock Improving antibody design with force-guided sampling in diffusion models.
\newblock \emph{arXiv preprint arXiv:2406.05832}, 2024.

\bibitem[Li et~al.(2025)Li, Cheng, Wu, Guo, Luo, Ren, Peng, and Ma]{li2024full}
Li, J., Cheng, C., Wu, Z., Guo, R., Luo, S., Ren, Z., Peng, J., and Ma, J.
\newblock Full-atom peptide design based on multi-modal flow matching.
\newblock In \emph{Proceedings of the 41st International Conference on Machine Learning}, ICML'24. JMLR.org, 2025.

\bibitem[Li et~al.(2024)Li, Lan, Shi, Zhu, Lu, Pu, Lu, and Zhang]{li2024delineating}
Li, M., Lan, X., Shi, X., Zhu, C., Lu, X., Pu, J., Lu, S., and Zhang, J.
\newblock Delineating the stepwise millisecond allosteric activation mechanism of the class c gpcr dimer mglu5.
\newblock \emph{Nature Communications}, 15\penalty0 (1):\penalty0 7519, 2024.

\bibitem[Lin et~al.(2025)Lin, Zhang, Zhao, Jiang, Wu, Liu, Huang, and Li]{lin2024ppflow}
Lin, H., Zhang, O., Zhao, H., Jiang, D., Wu, L., Liu, Z., Huang, Y., and Li, S.~Z.
\newblock Ppflow: target-aware peptide design with torsional flow matching.
\newblock In \emph{Proceedings of the 41st International Conference on Machine Learning}, ICML'24. JMLR.org, 2025.

\bibitem[Lin et~al.(2023)Lin, Akin, Rao, Hie, Zhu, Lu, Smetanin, Verkuil, Kabeli, Shmueli, et~al.]{lin2023evolutionary}
Lin, Z., Akin, H., Rao, R., Hie, B., Zhu, Z., Lu, W., Smetanin, N., Verkuil, R., Kabeli, O., Shmueli, Y., et~al.
\newblock Evolutionary-scale prediction of atomic-level protein structure with a language model.
\newblock \emph{Science}, 379\penalty0 (6637):\penalty0 1123--1130, 2023.

\bibitem[Lisanza et~al.(2024)Lisanza, Gershon, Tipps, Sims, Arnoldt, Hendel, Simma, Liu, Yase, Wu, et~al.]{lisanza2024multistate}
Lisanza, S.~L., Gershon, J.~M., Tipps, S.~W., Sims, J.~N., Arnoldt, L., Hendel, S.~J., Simma, M.~K., Liu, G., Yase, M., Wu, H., et~al.
\newblock Multistate and functional protein design using rosettafold sequence space diffusion.
\newblock \emph{Nature Biotechnology}, pp.\  1--11, 2024.

\bibitem[Liu et~al.(2024)Liu, Yang, Cao, Gao, Yang, Zhang, Zhu, and Wu]{liu2024cyclicpepedia}
Liu, L., Yang, L., Cao, S., Gao, Z., Yang, B., Zhang, G., Zhu, R., and Wu, D.
\newblock Cyclicpepedia: a knowledge base of natural and synthetic cyclic peptides.
\newblock \emph{Briefings in Bioinformatics}, 25\penalty0 (3):\penalty0 bbae190, 2024.

\bibitem[Loshchilov(2017)]{loshchilov2017decoupled}
Loshchilov, I.
\newblock Decoupled weight decay regularization.
\newblock \emph{arXiv preprint arXiv:1711.05101}, 2017.

\bibitem[Lu et~al.(2023)Lu, Chen, Chen, Su, Li, and Zhu]{lu2023contrastive}
Lu, C., Chen, H., Chen, J., Su, H., Li, C., and Zhu, J.
\newblock Contrastive energy prediction for exact energy-guided diffusion sampling in offline reinforcement learning.
\newblock In \emph{International Conference on Machine Learning}, pp.\  22825--22855. PMLR, 2023.

\bibitem[Lu et~al.(2022)Lu, Wu, Zhang, Rao, Li, and Zheng]{lu2022tankbind}
Lu, W., Wu, Q., Zhang, J., Rao, J., Li, C., and Zheng, S.
\newblock Tankbind: Trigonometry-aware neural networks for drug-protein binding structure prediction.
\newblock \emph{Advances in neural information processing systems}, 35:\penalty0 7236--7249, 2022.

\bibitem[Lu et~al.(2024)Lu, Zhang, Huang, Zhang, Jia, Wang, Shi, Li, Wolynes, and Zheng]{lu2024dynamicbind}
Lu, W., Zhang, J., Huang, W., Zhang, Z., Jia, X., Wang, Z., Shi, L., Li, C., Wolynes, P.~G., and Zheng, S.
\newblock Dynamicbind: Predicting ligand-specific protein-ligand complex structure with a deep equivariant generative model.
\newblock \emph{Nature Communications}, 15\penalty0 (1):\penalty0 1071, 2024.

\bibitem[Luo et~al.(2022)Luo, Su, Peng, Wang, Peng, and Ma]{luo2022antigen}
Luo, S., Su, Y., Peng, X., Wang, S., Peng, J., and Ma, J.
\newblock Antigen-specific antibody design and optimization with diffusion-based generative models for protein structures.
\newblock \emph{Advances in Neural Information Processing Systems}, 35:\penalty0 9754--9767, 2022.

\bibitem[Madani et~al.(2023)Madani, Krause, Greene, Subramanian, Mohr, Holton, Olmos, Xiong, Sun, Socher, et~al.]{madani2023large}
Madani, A., Krause, B., Greene, E.~R., Subramanian, S., Mohr, B.~P., Holton, J.~M., Olmos, J.~L., Xiong, C., Sun, Z.~Z., Socher, R., et~al.
\newblock Large language models generate functional protein sequences across diverse families.
\newblock \emph{Nature Biotechnology}, 41\penalty0 (8):\penalty0 1099--1106, 2023.

\bibitem[Maier et~al.(2015)Maier, Martinez, Kasavajhala, Wickstrom, Hauser, and Simmerling]{maier2015ff14sb}
Maier, J.~A., Martinez, C., Kasavajhala, K., Wickstrom, L., Hauser, K.~E., and Simmerling, C.
\newblock ff14sb: improving the accuracy of protein side chain and backbone parameters from ff99sb.
\newblock \emph{Journal of chemical theory and computation}, 11\penalty0 (8):\penalty0 3696--3713, 2015.

\bibitem[Makowski et~al.(2024)Makowski, Wang, Zupancic, Huang, Wu, Schardt, De~Groot, Elkins, Martin, and Tessier]{makowski2024optimization}
Makowski, E.~K., Wang, T., Zupancic, J.~M., Huang, J., Wu, L., Schardt, J.~S., De~Groot, A.~S., Elkins, S.~L., Martin, W.~D., and Tessier, P.~M.
\newblock Optimization of therapeutic antibodies for reduced self-association and non-specific binding via interpretable machine learning.
\newblock \emph{Nature biomedical engineering}, 8\penalty0 (1):\penalty0 45--56, 2024.

\bibitem[Mao et~al.(2024)Mao, Zhu, Sun, Shen, Wu, Chen, and Shen]{mao2024de}
Mao, W., Zhu, M., Sun, Z., Shen, S., Wu, L.~Y., Chen, H., and Shen, C.
\newblock De novo protein design using geometric vector field networks.
\newblock In \emph{The Twelfth International Conference on Learning Representations}, 2024.
\newblock URL \url{https://openreview.net/forum?id=9UIGyJJpay}.

\bibitem[Martinkus et~al.(2023)Martinkus, Ludwiczak, LIANG, Lafrance-Vanasse, Hotzel, Rajpal, Wu, Cho, Bonneau, Gligorijevic, and Loukas]{martinkus2023abdiffuser}
Martinkus, K., Ludwiczak, J., LIANG, W.-C., Lafrance-Vanasse, J., Hotzel, I., Rajpal, A., Wu, Y., Cho, K., Bonneau, R., Gligorijevic, V., and Loukas, A.
\newblock Abdiffuser: full-atom generation of in-vitro functioning antibodies.
\newblock In \emph{Thirty-seventh Conference on Neural Information Processing Systems}, 2023.
\newblock URL \url{https://openreview.net/forum?id=7GyYpomkEa}.

\bibitem[Martins et~al.(2023)Martins, Mariano, Carvalho, Bastos, Moraes, Paix{\~a}o, and Cardoso~de Melo-Minardi]{martins2023propedia}
Martins, P., Mariano, D., Carvalho, F.~C., Bastos, L.~L., Moraes, L., Paix{\~a}o, V., and Cardoso~de Melo-Minardi, R.
\newblock Propedia v2. 3: A novel representation approach for the peptide-protein interaction database using graph-based structural signatures.
\newblock \emph{Frontiers in Bioinformatics}, 3:\penalty0 1103103, 2023.

\bibitem[Merz et~al.(2024)Merz, Habeshian, Li, David, Nielsen, Ji, Il~Khwildy, Duany~Benitez, Phothirath, and Heinis]{merz2024novo}
Merz, M.~L., Habeshian, S., Li, B., David, J.-A.~G., Nielsen, A.~L., Ji, X., Il~Khwildy, K., Duany~Benitez, M.~M., Phothirath, P., and Heinis, C.
\newblock De novo development of small cyclic peptides that are orally bioavailable.
\newblock \emph{Nature Chemical Biology}, 20\penalty0 (5):\penalty0 624--633, 2024.

\bibitem[Peacock \& Suga(2021)Peacock and Suga]{peacock2021discovery}
Peacock, H. and Suga, H.
\newblock Discovery of de novo macrocyclic peptides by messenger rna display.
\newblock \emph{Trends in Pharmacological Sciences}, 42\penalty0 (5):\penalty0 385--397, 2021.

\bibitem[Pei et~al.(2024)Pei, Gao, Wu, Zhu, Xia, Xie, Qin, He, Liu, and Yan]{pei2024fabind}
Pei, Q., Gao, K., Wu, L., Zhu, J., Xia, Y., Xie, S., Qin, T., He, K., Liu, T.-Y., and Yan, R.
\newblock Fabind: Fast and accurate protein-ligand binding.
\newblock \emph{Advances in Neural Information Processing Systems}, 36, 2024.

\bibitem[Qian \& Zhou(2006)Qian and Zhou]{qian2006set}
Qian, C. and Zhou, M.~M.
\newblock Set domain protein lysine methyltransferases: Structure, specificity and catalysis.
\newblock \emph{Cellular and molecular life sciences CMLS}, 63:\penalty0 2755--2763, 2006.

\bibitem[Qiao et~al.(2024)Qiao, Nie, Vahdat, Miller~III, and Anandkumar]{qiao2024state}
Qiao, Z., Nie, W., Vahdat, A., Miller~III, T.~F., and Anandkumar, A.
\newblock State-specific protein--ligand complex structure prediction with a multiscale deep generative model.
\newblock \emph{Nature Machine Intelligence}, 6\penalty0 (2):\penalty0 195--208, 2024.

\bibitem[Rettie et~al.(2024)Rettie, Juergens, Adebomi, Bueso, Zhao, Leveille, Liu, Bera, Wilms, {\"U}ffing, et~al.]{rettie2024accurate}
Rettie, S., Juergens, D., Adebomi, V., Bueso, Y.~F., Zhao, Q., Leveille, A., Liu, A., Bera, A., Wilms, J., {\"U}ffing, A., et~al.
\newblock Accurate de novo design of high-affinity protein binding macrocycles using deep learning.
\newblock \emph{bioRxiv}, pp.\  2024--11, 2024.

\bibitem[Rives et~al.(2021)Rives, Meier, Sercu, Goyal, Lin, Liu, Guo, Ott, Zitnick, Ma, et~al.]{rives2021biological}
Rives, A., Meier, J., Sercu, T., Goyal, S., Lin, Z., Liu, J., Guo, D., Ott, M., Zitnick, C.~L., Ma, J., et~al.
\newblock Biological structure and function emerge from scaling unsupervised learning to 250 million protein sequences.
\newblock \emph{Proceedings of the National Academy of Sciences}, 118\penalty0 (15):\penalty0 e2016239118, 2021.

\bibitem[Ryckaert et~al.(1977)Ryckaert, Ciccotti, and Berendsen]{ryckaert1977numerical}
Ryckaert, J.-P., Ciccotti, G., and Berendsen, H.~J.
\newblock Numerical integration of the cartesian equations of motion of a system with constraints: molecular dynamics of n-alkanes.
\newblock \emph{Journal of computational physics}, 23\penalty0 (3):\penalty0 327--341, 1977.

\bibitem[Salomon-Ferrer et~al.(2013)Salomon-Ferrer, Gotz, Poole, Le~Grand, and Walker]{salomon2013routine}
Salomon-Ferrer, R., Gotz, A.~W., Poole, D., Le~Grand, S., and Walker, R.~C.
\newblock Routine microsecond molecular dynamics simulations with amber on gpus. 2. explicit solvent particle mesh ewald.
\newblock \emph{Journal of chemical theory and computation}, 9\penalty0 (9):\penalty0 3878--3888, 2013.

\bibitem[S{\"a}rkk{\"a} \& Solin(2019)S{\"a}rkk{\"a} and Solin]{sarkka2019applied}
S{\"a}rkk{\"a}, S. and Solin, A.
\newblock \emph{Applied stochastic differential equations}, volume~10.
\newblock Cambridge University Press, 2019.

\bibitem[Satorras et~al.(2021)Satorras, Hoogeboom, and Welling]{satorras2021n}
Satorras, V.~G., Hoogeboom, E., and Welling, M.
\newblock E (n) equivariant graph neural networks.
\newblock In \emph{International conference on machine learning}, pp.\  9323--9332. PMLR, 2021.

\bibitem[Sharma et~al.(2023)Sharma, Sharma, Sharma, and Jain]{sharma2023peptide}
Sharma, K., Sharma, K.~K., Sharma, A., and Jain, R.
\newblock Peptide-based drug discovery: Current status and recent advances.
\newblock \emph{Drug Discovery Today}, 28\penalty0 (2):\penalty0 103464, 2023.

\bibitem[Song \& Ermon(2019)Song and Ermon]{song2019generative}
Song, Y. and Ermon, S.
\newblock Generative modeling by estimating gradients of the data distribution.
\newblock \emph{Advances in neural information processing systems}, 32, 2019.

\bibitem[Song et~al.(2021)Song, Sohl-Dickstein, Kingma, Kumar, Ermon, and Poole]{song2021scorebased}
Song, Y., Sohl-Dickstein, J., Kingma, D.~P., Kumar, A., Ermon, S., and Poole, B.
\newblock Score-based generative modeling through stochastic differential equations.
\newblock In \emph{International Conference on Learning Representations}, 2021.
\newblock URL \url{https://openreview.net/forum?id=PxTIG12RRHS}.

\bibitem[St{\"a}rk et~al.(2022)St{\"a}rk, Ganea, Pattanaik, Barzilay, and Jaakkola]{stark2022equibind}
St{\"a}rk, H., Ganea, O., Pattanaik, L., Barzilay, R., and Jaakkola, T.
\newblock Equibind: Geometric deep learning for drug binding structure prediction.
\newblock In \emph{International conference on machine learning}, pp.\  20503--20521. PMLR, 2022.

\bibitem[Stark et~al.(2023)Stark, Jing, Barzilay, and Jaakkola]{stark2023harmonic}
Stark, H., Jing, B., Barzilay, R., and Jaakkola, T.
\newblock Harmonic prior self-conditioned flow matching for multi-ligand docking and binding site design.
\newblock In \emph{NeurIPS 2023 AI for Science Workshop}, 2023.

\bibitem[Tang et~al.(2025)Tang, Zhang, and Chatterjee]{tang2025peptune}
Tang, S., Zhang, Y., and Chatterjee, P.
\newblock Peptune: De novo generation of therapeutic peptides with multi-objective-guided discrete diffusion.
\newblock \emph{ArXiv}, pp.\  arXiv--2412, 2025.

\bibitem[Tsaban et~al.(2022)Tsaban, Varga, Avraham, Ben-Aharon, Khramushin, and Schueler-Furman]{tsaban2022harnessing}
Tsaban, T., Varga, J.~K., Avraham, O., Ben-Aharon, Z., Khramushin, A., and Schueler-Furman, O.
\newblock Harnessing protein folding neural networks for peptide--protein docking.
\newblock \emph{Nature communications}, 13\penalty0 (1):\penalty0 176, 2022.

\bibitem[Tsomaia(2015)]{tsomaia2015peptide}
Tsomaia, N.
\newblock Peptide therapeutics: targeting the undruggable space.
\newblock \emph{European journal of medicinal chemistry}, 94:\penalty0 459--470, 2015.

\bibitem[van Gelder et~al.(2022)van Gelder, Lerma, Engelke, and Huizinga]{van2022voclosporin}
van Gelder, T., Lerma, E., Engelke, K., and Huizinga, R.~B.
\newblock Voclosporin: a novel calcineurin inhibitor for the treatment of lupus nephritis.
\newblock \emph{Expert review of clinical pharmacology}, 15\penalty0 (5):\penalty0 515--529, 2022.

\bibitem[Vinogradov et~al.(2019)Vinogradov, Yin, and Suga]{vinogradov2019macrocyclic}
Vinogradov, A.~A., Yin, Y., and Suga, H.
\newblock Macrocyclic peptides as drug candidates: recent progress and remaining challenges.
\newblock \emph{Journal of the American Chemical Society}, 141\penalty0 (10):\penalty0 4167--4181, 2019.

\bibitem[Wang et~al.(2019)Wang, Sun, Wang, Wang, Liu, Zhang, and Hou]{wang2019end}
Wang, E., Sun, H., Wang, J., Wang, Z., Liu, H., Zhang, J.~Z., and Hou, T.
\newblock End-point binding free energy calculation with mm/pbsa and mm/gbsa: strategies and applications in drug design.
\newblock \emph{Chemical reviews}, 119\penalty0 (16):\penalty0 9478--9508, 2019.

\bibitem[Wang et~al.(2024{\natexlab{a}})Wang, Wang, Feng, Zhang, and Lai]{wang2024target}
Wang, F., Wang, Y., Feng, L., Zhang, C., and Lai, L.
\newblock Target-specific de novo peptide binder design with diffpepbuilder.
\newblock \emph{Journal of Chemical Information and Modeling}, 2024{\natexlab{a}}.

\bibitem[Wang et~al.(2005)Wang, Fang, Lu, Yang, and Wang]{wang2005pdbbind}
Wang, R., Fang, X., Lu, Y., Yang, C.-Y., and Wang, S.
\newblock The pdbbind database: methodologies and updates.
\newblock \emph{Journal of medicinal chemistry}, 48\penalty0 (12):\penalty0 4111--4119, 2005.

\bibitem[Wang et~al.(2024{\natexlab{b}})Wang, Zheng, Fei, Xue, Huang, and Gu]{wang2024dplm}
Wang, X., Zheng, Z., Fei, Y., Xue, D., Huang, S., and Gu, Q.
\newblock Diffusion language models are versatile protein learners.
\newblock In \emph{International conference on machine learning}, 2024{\natexlab{b}}.

\bibitem[Wang et~al.(2025)Wang, Zheng, Ye, Xue, Huang, and Gu]{wang2025dplm2}
Wang, X., Zheng, Z., Ye, F., Xue, D., Huang, S., and Gu, Q.
\newblock Dplm-2: A multimodal diffusion protein language model.
\newblock In \emph{International Conference on Learning Representations}, 2025.

\bibitem[Watson et~al.(2023)Watson, Juergens, Bennett, Trippe, Yim, Eisenach, Ahern, Borst, Ragotte, Milles, et~al.]{watson2023novo}
Watson, J.~L., Juergens, D., Bennett, N.~R., Trippe, B.~L., Yim, J., Eisenach, H.~E., Ahern, W., Borst, A.~J., Ragotte, R.~J., Milles, L.~F., et~al.
\newblock De novo design of protein structure and function with rfdiffusion.
\newblock \emph{Nature}, 620\penalty0 (7976):\penalty0 1089--1100, 2023.

\bibitem[Wei et~al.(2024)Wei, Wang, Peng, and Yang]{wei2024q}
Wei, H., Wang, W., Peng, Z., and Yang, J.
\newblock Q-biolip: A comprehensive resource for quaternary structure-based protein--ligand interactions.
\newblock \emph{Genomics, Proteomics \& Bioinformatics}, 22\penalty0 (1), 2024.

\bibitem[Wen et~al.(2019)Wen, He, Tao, and Huang]{wen2019pepbdb}
Wen, Z., He, J., Tao, H., and Huang, S.-Y.
\newblock Pepbdb: a comprehensive structural database of biological peptide--protein interactions.
\newblock \emph{Bioinformatics}, 35\penalty0 (1):\penalty0 175--177, 2019.

\bibitem[Weng et~al.(2020)Weng, Gao, Wang, Wang, Hu, Yao, Cao, and Hou]{doi:10.1021/acs.jctc.9b01208}
Weng, G., Gao, J., Wang, Z., Wang, E., Hu, X., Yao, X., Cao, D., and Hou, T.
\newblock Comprehensive evaluation of fourteen docking programs on protein–peptide complexes.
\newblock \emph{Journal of Chemical Theory and Computation}, 16\penalty0 (6):\penalty0 3959--3969, 2020.
\newblock \doi{10.1021/acs.jctc.9b01208}.
\newblock URL \url{https://doi.org/10.1021/acs.jctc.9b01208}.
\newblock PMID: 32324992.

\bibitem[Xie et~al.(2023)Xie, Valiente, and Kim]{xie2023helixgan}
Xie, X., Valiente, P.~A., and Kim, P.~M.
\newblock Helixgan a deep-learning methodology for conditional de novo design of $\alpha$-helix structures.
\newblock \emph{Bioinformatics}, 39\penalty0 (1):\penalty0 btad036, 2023.

\bibitem[Xie et~al.(2024)Xie, Valiente, Kim, and Kim]{xie2024helixdiff}
Xie, X., Valiente, P.~A., Kim, J., and Kim, P.~M.
\newblock Helixdiff, a score-based diffusion model for generating all-atom $\alpha$-helical structures.
\newblock \emph{ACS Central Science}, 10\penalty0 (5):\penalty0 1001--1011, 2024.

\bibitem[Yang et~al.(2021)Yang, Wang, Zhou, and Zhang]{yang2021histone}
Yang, C., Wang, K., Zhou, Y., and Zhang, S.-L.
\newblock Histone lysine methyltransferase set8 is a novel therapeutic target for cancer treatment.
\newblock \emph{Drug Discovery Today}, 26\penalty0 (10):\penalty0 2423--2430, 2021.

\bibitem[Ye et~al.(2024)Ye, Zheng, Xue, Shen, Wang, Ma, Wang, Wang, Zhou, and Gu]{ye2024proteinbench}
Ye, F., Zheng, Z., Xue, D., Shen, Y., Wang, L., Ma, Y., Wang, Y., Wang, X., Zhou, X., and Gu, Q.
\newblock Proteinbench: A holistic evaluation of protein foundation models.
\newblock \emph{arXiv preprint arXiv:2409.06744}, 2024.

\bibitem[Yi et~al.(2024)Yi, Zhou, Shen, Li{\`o}, and Wang]{yi2024graph}
Yi, K., Zhou, B., Shen, Y., Li{\`o}, P., and Wang, Y.
\newblock Graph denoising diffusion for inverse protein folding.
\newblock \emph{Advances in Neural Information Processing Systems}, 36, 2024.

\bibitem[Yim et~al.(2023)Yim, Trippe, De~Bortoli, Mathieu, Doucet, Barzilay, and Jaakkola]{yim2023se}
Yim, J., Trippe, B.~L., De~Bortoli, V., Mathieu, E., Doucet, A., Barzilay, R., and Jaakkola, T.
\newblock Se (3) diffusion model with application to protein backbone generation.
\newblock In \emph{Proceedings of the 40th International Conference on Machine Learning}, pp.\  40001--40039, 2023.

\bibitem[Yim et~al.(2024)Yim, Campbell, Mathieu, Foong, Gastegger, Jiménez-Luna, Lewis, Satorras, Veeling, Noé, Barzilay, and Jaakkola]{yim2024improvedmotifscaffoldingse3flow}
Yim, J., Campbell, A., Mathieu, E., Foong, A. Y.~K., Gastegger, M., Jiménez-Luna, J., Lewis, S., Satorras, V.~G., Veeling, B.~S., Noé, F., Barzilay, R., and Jaakkola, T.~S.
\newblock Improved motif-scaffolding with se(3) flow matching, 2024.
\newblock URL \url{https://arxiv.org/abs/2401.04082}.

\bibitem[Zhang et~al.(2013)Zhang, Tanaka, Yan, Li, Peng, Jiang, Yang, Barton, Wen, and Shi]{zhang2013regulation}
Zhang, X., Tanaka, K., Yan, J., Li, J., Peng, D., Jiang, Y., Yang, Z., Barton, M.~C., Wen, H., and Shi, X.
\newblock Regulation of estrogen receptor $\alpha$ by histone methyltransferase smyd2-mediated protein methylation.
\newblock \emph{Proceedings of the National Academy of Sciences}, 110\penalty0 (43):\penalty0 17284--17289, 2013.

\bibitem[Zhang \& Skolnick(2005)Zhang and Skolnick]{zhang2005tm}
Zhang, Y. and Skolnick, J.
\newblock Tm-align: a protein structure alignment algorithm based on the tm-score.
\newblock \emph{Nucleic acids research}, 33\penalty0 (7):\penalty0 2302--2309, 2005.

\bibitem[Zheng et~al.(2022)Zheng, Zhang, and Rao]{zheng2022protein}
Zheng, Q., Zhang, W., and Rao, G.-W.
\newblock Protein lysine methyltransferase smyd2: a promising small molecule target for cancer therapy.
\newblock \emph{Journal of Medicinal Chemistry}, 65\penalty0 (15):\penalty0 10119--10132, 2022.

\bibitem[Zheng et~al.(2023)Zheng, Deng, Xue, Zhou, Ye, and Gu]{zheng2023structure}
Zheng, Z., Deng, Y., Xue, D., Zhou, Y., Ye, F., and Gu, Q.
\newblock Structure-informed language models are protein designers.
\newblock In \emph{International conference on machine learning}, pp.\  42317--42338. PMLR, 2023.

\bibitem[Zhou et~al.(2023)Zhou, Gao, Ding, Zheng, Xu, Wei, Zhang, and Ke]{zhouuni}
Zhou, G., Gao, Z., Ding, Q., Zheng, H., Xu, H., Wei, Z., Zhang, L., and Ke, G.
\newblock Uni-mol: A universal 3d molecular representation learning framework.
\newblock In \emph{The Eleventh International Conference on Learning Representations}, 2023.

\bibitem[Zhou et~al.(2024{\natexlab{a}})Zhou, Cheng, Yang, Bao, Wang, and Gu]{zhou2024decompopt}
Zhou, X., Cheng, X., Yang, Y., Bao, Y., Wang, L., and Gu, Q.
\newblock Decompopt: Controllable and decomposed diffusion models for structure-based molecular optimization.
\newblock \emph{arXiv preprint arXiv:2403.13829}, 2024{\natexlab{a}}.

\bibitem[Zhou et~al.(2024{\natexlab{b}})Zhou, Guan, Zhang, Peng, Wang, and Ma]{zhou2024reprogramming}
Zhou, X., Guan, J., Zhang, Y., Peng, X., Wang, L., and Ma, J.
\newblock Reprogramming pretrained target-specific diffusion models for dual-target drug design.
\newblock \emph{Advances in Neural Information Processing Systems}, 37:\penalty0 87255--87281, 2024{\natexlab{b}}.

\bibitem[Zhou et~al.(2024{\natexlab{c}})Zhou, Wang, and Zhou]{zhou2024stabilizing}
Zhou, X., Wang, L., and Zhou, Y.
\newblock Stabilizing policy gradients for stochastic differential equations via consistency with perturbation process.
\newblock \emph{arXiv preprint arXiv:2403.04154}, 2024{\natexlab{c}}.

\bibitem[Zhou et~al.(2024{\natexlab{d}})Zhou, Xue, Chen, Zheng, Wang, and Gu]{zhou2024antigen}
Zhou, X., Xue, D., Chen, R., Zheng, Z., Wang, L., and Gu, Q.
\newblock Antigen-specific antibody design via direct energy-based preference optimization.
\newblock \emph{Advances in Neural Information Processing Systems}, 37:\penalty0 120861--120891, 2024{\natexlab{d}}.

\bibitem[Zhou et~al.(2025)Zhou, Xiao, Lin, He, Guan, Wang, Liu, Zhou, Wang, and Ma]{zhou2025integrating}
Zhou, X., Xiao, Y., Lin, H., He, X., Guan, J., Wang, Y., Liu, Q., Zhou, F., Wang, L., and Ma, J.
\newblock Integrating protein dynamics into structure-based drug design via full-atom stochastic flows.
\newblock \emph{arXiv preprint arXiv:2503.03989}, 2025.

\bibitem[Zorzi et~al.(2017)Zorzi, Deyle, and Heinis]{zorzi2017cyclicthe}
Zorzi, A., Deyle, K., and Heinis, C.
\newblock Cyclic peptide therapeutics: past, present and future.
\newblock \emph{Current opinion in chemical biology}, 38:\penalty0 24--29, 2017.

\bibitem[Zotchev et~al.(2006)Zotchev, Stepanchikova, Sergeyko, Sobolev, Filimonov, and Poroikov]{zotchev2006rational}
Zotchev, S.~B., Stepanchikova, A.~V., Sergeyko, A.~P., Sobolev, B.~N., Filimonov, D.~A., and Poroikov, V.~V.
\newblock Rational design of macrolides by virtual screening of combinatorial libraries generated through in silico manipulation of polyketide synthases.
\newblock \emph{Journal of medicinal chemistry}, 49\penalty0 (6):\penalty0 2077--2087, 2006.

\end{thebibliography}
\bibliographystyle{icml2025}

\newpage
\appendix
\onecolumn

\section{Introduction of Cyclic Peptides}

Peptides have shown promising capability as therapeutics for protein targets where small molecules struggle to bind, due to their ability to modulate protein-protein interactions \citep{hosseinzadeh2021anchor, zorzi2017cyclicthe}. However, traditional linear peptides are usually polar due to exposed acids and amines in terminals, limiting their membrane permeability and proteolytic stability, and also restricts their administration options in drug development~\citep{buckton2021cyclicas, merz2024novo}.

\begin{figure}[h]
    \centering
    \includegraphics[width=0.68\linewidth]{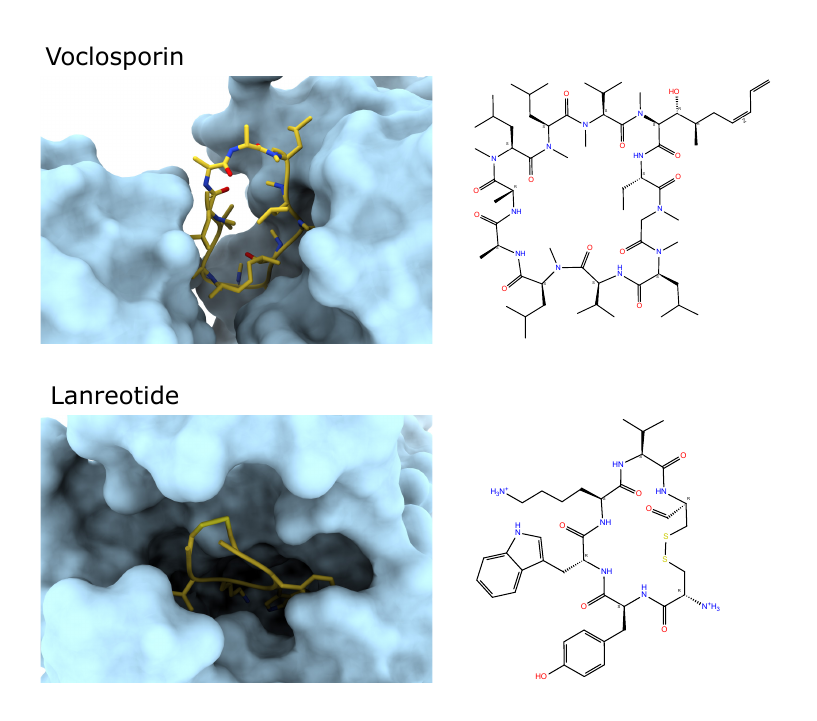}
    \caption{Two examples of cyclic peptide drugs. Voclosporin \citep{van2022voclosporin}, an analog of ciclosporin, is an immunosuppressant used to treat lupus nephritis. Lanreotide \citep{caplin2014lanreotide}, an analog of somatostatin, is an oncology drug that inhibits growth hormone release and is used to manage carcinoid syndrome.}
    \label{fig:cyclic_peptide_drug}
\end{figure}

The peptide cyclization is capable of enhancing the conformational stability, increasing binding affinity and specificity for targets \citep{zorzi2017cyclicthe}. This has fueled the growing interest in cyclic peptide research as efficient therapeutics. \cref{fig:cyclic_peptide_drug} shows two cyclic peptide drugs approved in recent 20 years. According to \citet{sharma2023peptide, fang2024recent, costa2023cyclic}, cyclic peptides can be classified by their cyclization strategies: head-to-tail cyclization (between N- and C-termini); side-to-side cyclization (between two side chains, including peptide stapling, which stabilizes $\alpha$-helical structures) \citep{vinogradov2019macrocyclic}; head-to-side and side-to-tail cyclization (between a terminal and side chain); and polycyclization (with multiple cycles).

\begin{figure}[h]
    \centering
    \includegraphics[width=0.98\linewidth]{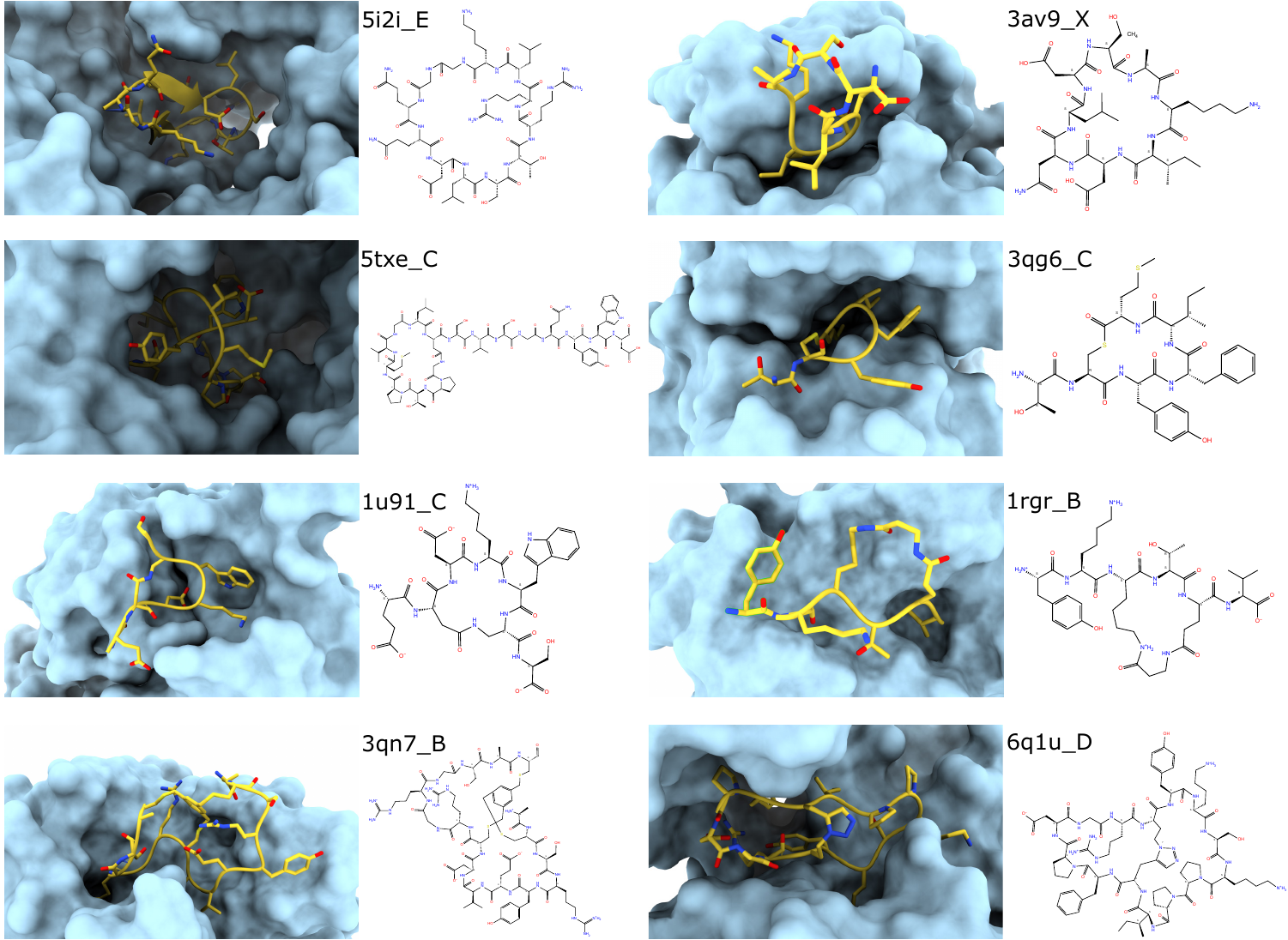}
    \caption{Examples of 3D structures of cyclic peptides and their chemical graphs. The cyclization structures are highlighted in pink. The corresponding cyclization types are: 5I2I\_E: head-to-tail; 3AV9\_X: head-to-tail; 5TXE\_C: head-to-side; 3QG6\_C: side-to-tail; 1U91\_C: side-to-side; 1RGR\_B: side-to-side (stapling); 3QN7\_B and 6Q1U\_D: polycyclization.}
    \label{fig:cyclic_peptide_exp}
\end{figure}

\begin{figure}[h]
    \centering
    \includegraphics[width=0.98\linewidth]{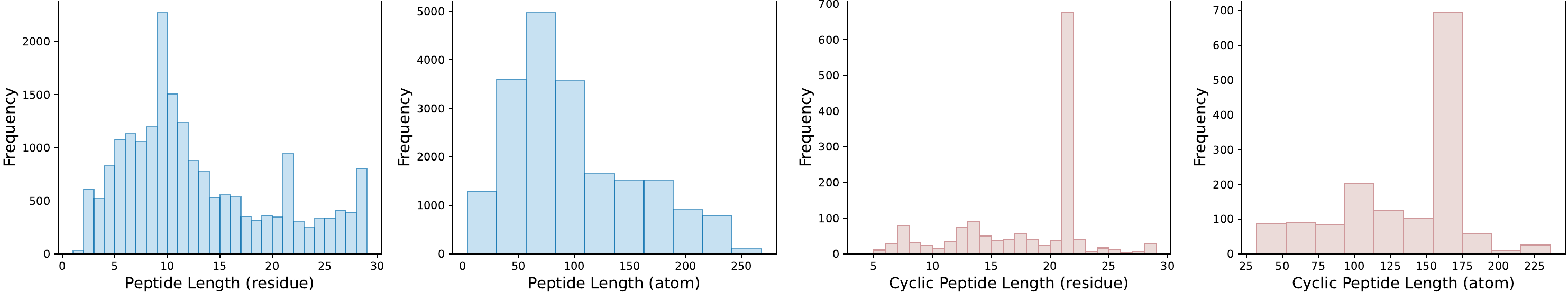}
    \caption{Statistics on peptide and cyclic peptide lengths in our dataset.}
    \label{fig:dataset_res_atom_num}
\end{figure}

\begin{figure}[h]
    \centering
    \includegraphics[width=0.98\linewidth]{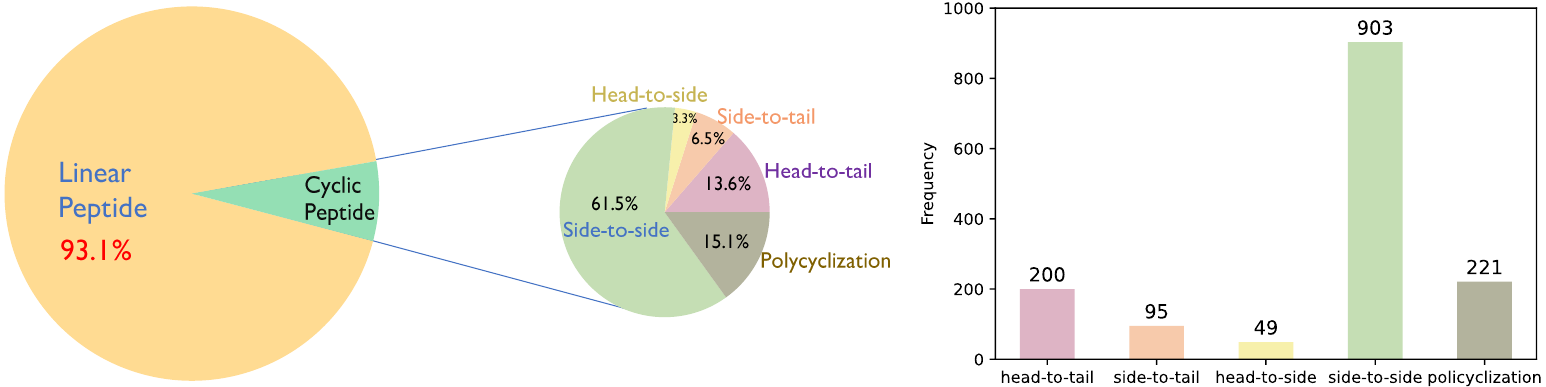}
    \caption{Analysis of cyclic peptide distribution and the proportions of the five cyclization types.}
    \label{fig:cyclic_peptide_data}
\end{figure}

\section{Extend Related Work}

\textbf{Protein-ligand Docking.} Protein-ligand docking aims to predict the conformation of protein-ligand complexes, playing a crucial role in drug discovery and design. Compared to traditional score-based or template-based docking methods \citep{eberhardt2021autodock,friesner2004glide}, deep learning-based approaches are faster while achieving comparable accuracy. A common strategy involves leveraging equivariant or invariant graph neural networks to model both protein and ligand and predict ligand atom positions \citep{lu2022tankbind,stark2022equibind,pei2024fabind,qiao2024state}. Other approaches, such as Uni-Mol~\citep{zhouuni} and RoseTTAFold All-Atom~\citep{krishna2024generalized}, use positional encoding and attention layers for atom-level representations and predictions. Unlike structure prediction models, Diffdock~\citep{corsodiffdock} introduces the generative docking framework that views docking as a conditional generation problem, and uses a diffusion model for end-to-end ligand generation. SurfDock \citep{cao2024surfdock} further incorporates protein surface graphs to refine docking predictions. More recent studies have emphasized flexible docking models, which consider protein conformational variability in docking process. DynamicBind \citep{lu2024dynamicbind} and ReDock \citep{huang2024redock} extend the generative docking paradigm by incorporating diffusion-based modeling of protein structures, enabling joint optimization of both protein and ligand poses. \citet{guan2025group} proposed a novel molecular docking framework that simultaneously considers multiple ligands docking to a protein, enhacing molecular docking accuracy.

\textbf{Protein Sequence Design.} Designing protein sequences that fold into desired structures, known as inverse folding, is a fundamental task in protein engineering. Deep learning methods can effectively predict protein sequences given structural priors. These models generally follow a two-stage paradigm comprising a structure encoding module and a sequence prediction module. Several models, including GVP \citep{jing2021learning}, ESM-IF \citep{hsu2022learning}, ProteinMPNN \citep{dauparas2022robust}, and \citet{ingraham2019generative}, utilize graph neural networks (GNNs) for structural encoding and employ autoregressive decoding for sequence prediction. PiFold \citep{gaopifold} adopts a similar encoding strategy but directly classifies amino acids for each node. GRADE-IF~\citep{yi2024graph} formulates the inverse folding problem as learning the conditional distribution of protein sequence given the backbone structure, and employs a discrete diffusion model on protein graph to predict the amino acid type on each node. Besides graph-based approaches, Protein Language Models (PLMs) \citep{lin2023evolutionary,rives2021biological} offer an alternative strategy for inverse folding. \citet{zheng2023structure} integrates PLMs as sequence decoders for encoded protein graphs, while \citet{gao2023kw} and \citet{mao2024de} incorporate ESM-2 \citep{lin2023evolutionary} embeddings to refine sequence predictions. ProGen \citep{madani2023large} shows an autoregressive language model trained on massive protein sequence data can utilize property tags for controllable sequence generation. 
\citet{wang2024dplm} demonstrate that discrete diffusion serves as a more principled probabilistic framework for large-scale protein language modeling. The resulting diffusion protein language model (DPLM) excels in both sequence generation and representation learning. DPLM-2~\citep{wang2025dplm2} further extends this discrete diffusion-based paradigm by incorporating tokenized 3D structures thereafter, enabling structure-sequence co-generation as well as any-to-any conditional generation with multimodal generative protein language models.

\textbf{Cyclic Peptide Design.} Although computational cyclic peptide design is an innovative research area, there are already recent studies on cyclic peptide design that are significantly different from our approach \citep{tang2025peptune,rettie2024accurate}. 
The approach proposed by \citet{tang2025peptune} is a ligand-based drug design (LBDD) method that models the sequence of cyclic peptides using discrete diffusion, optimized by multiple reward functions. It does not explicitly incorporate the 3D structure of target proteins, whereas our structure-based drug design (SBDD) method directly designs ligands based on 3D target structures. \citet{rettie2024accurate} uses modified RoseTTAFold \citep{baek2021accurate} and RFdiffusion \citep{watson2023novo} with cyclic relative positional encoding to generate macrocyclic backbones.

\section{Protein all-atom representation}
\textbf{Atom14 representation.} The atom14 encoding efficiently represents residues using 14 columns to capture atom content, with 14 being the maximum number of heavy atoms in the 20 canonical amino acids (e.g., Tryptophan). Empty strings are padded in atom14 representation when a residue contains fewer than 14 atoms. 

\textbf{Atom37 representation.} The atom37 encoding is a fixed-dimension representation where each of the 37 heavy atom types in canonical amino acids is assigned a unique position in an array. For amino acids missing certain atoms, padding is used to maintain a consistent length. The atom37 format includes the following atoms: [`N', `CA', `C', `CB', `O', `CG', `CG1', `CG2', `OG', `OG1', `SG', `CD', `CD1', `CD2', `ND1', `ND2', `OD1', `OD2', `SD', `CE', `CE1', `CE2', `CE3', `NE', `NE1', `NE2', `OE1', `OE2', `CH2', `NH1', `NH2', `OH', `CZ', `CZ2', `CZ3', `NZ', `OXT'].

\textbf{Atom73 representation.} The atom73 representation, introduced by \citet{chu2024all}, indexes the `N', `CA', `C', `CB', and `O' atoms, then independently encodes each amino acid's side chains. Backbone atoms share the same position, while side-chain atoms are assigned to specific residue types. For example, the `NE2' atom in the `GLN' residue is recorded as `Q-NE2' (where Q is the one-letter code for `GLN') and occupies a unique column in the atom73 encoding.

\section{Peptide Dataset}
\label{app:peptide_dataset}

\textbf{CyclicPepedia.} The CyclicPepedia dataset \citep{liu2024cyclicpepedia} contains 9,744 cyclic peptides from multiple sources. It serves as a knowledge database with information on categorization, structural characteristics, pharmacokinetics, physicochemical properties, patented drug applications, and key publications. However, most peptide conformations are generated by RDKit, with only 1,325 cyclic peptides having original 3D structures and 61 containing complex structures. The remaining data only include fingerprints, limiting its application in structure-based drug design.

\textbf{PPBench2024.} PPBench2024 \citep{lin2024ppflow} is a protein-peptide binding dataset sourced from \citet{burley2023rcsb}, \citet{martins2023propedia}, and \citet{doi:10.1021/acs.jctc.9b01208}. It selects complexes with more than two chains and an interaction distance under 5 Å. Complexes with peptides shorter than 30 amino acids are then filtered. Non-peptide molecules and peptides with unusual bond lengths or non-amino acid functional groups are also excluded. This results in a total of 15,593 protein-peptide pairs.

\textbf{PepBench.} PepBench is a curated benchmarking dataset designed to train and evaluate protein-peptide binding models in  PepGLAD~\citep{kong2024fullatom}. The training data is sourced from \citep{berman2000protein}, while the test set is adopted from \citet{tsaban2022harnessing}. To ensure quality, peptides are restricted to lengths between 4 and 25 residues, and receptors to more than 30 residues. Complexes with over 90\% sequence similarity are excluded to reduce redundancy. Furthermore, clustering at 40\% sequence identity is applied to separate training and test data, removing any training complexes that share clusters with the test set. The final dataset comprises 4,157 training complexes, 114 for validation, and 93 for testing. 

\textbf{PepFlow.} The dataset introduced by \citet{li2024full} combines data from \citet{wen2019pepbdb} and \citet{wei2024q}, resulting in 8,365 protein-peptide complex structures. Peptides are filtered to include those with lengths ranging from 3 to 25 amino acids and are clustered at 40\% sequence identity. From these clusters, 158 structures are selected for the test set, ensuring each cluster has 10 to 50 members. The remaining data is used for training and validation.

\textbf{Our curation.} Our curation has been roughly introduced in \cref{subsec:experimental_setup}. A unique requirement of our method is converting Structured Data Files (SDF files) to PDB files, as the chemical bond information in PDB format is often incomplete. We found that RDKit~\footnote{\url{https://www.rdkit.org/}} does not always accurately produce chemical bonds during conversion, as it determines bond existence and type based on pairwise atom types and distances. Therefore, we leverage the Chemical Component Dictionary (CCD)~\footnote{\url{https://www.wwpdb.org/data/ccd}}, which describes all residue and small molecule components found in PDB entries, to collect all intra-residue bond information. We use default peptide bonds to connect canonical residues with continuous residue indices and the bonds stored in the CONECT information to recover bonds between non-canonical residues and other residues.

\section{Routed Sampling}
\label{app:routed_sampling}
The general idea of the proposed routed sampling is that the sequence and structure are alternately updated in the generative process. We present its details in \cref{alg:routed_sampling}. 
In this subsection, we introduce some notations without rigorous definitions but maintain clarity, as we provide detailed comments after the lines in the algorithm.
The sampling algorithm is inspired by \citet{chu2024all}. However, we innovatively introduce a dynamic chemical graph to make the all-atom peptide design compatible with cyclic structures, especially non-canonical covalent bonds and residues.
Several critical functions used in \cref{alg:routed_sampling} will be discussed as follows:

\begin{compactitem}
    \item ``Extract'' and ``Cache'': As introduced in \cref{subsec:sampling}, due to cyclization, we can categorize atoms into two types: constrained atoms and free-residue atoms. We maintain the atom73 states and individual atom states for these types, respectively. Two functions are employed to extract and store atom coordinates based on masks determined by residue types or constrained atom indices.
    \item ``Assemble'': Each time the residue types are updated, the corresponding side-chain chemical graphs are likewise updated. Consequently, we reassemble each residue's chemical graph along with the cyclization chemical graph into a complete peptide chemical graph to serve as input for the models.
    \item ``Supgraph'': As introduced in \cref{subsec:seqmodel}, to avoid residue type information leakage from the side-chain chemical graphs, we remove the side-chain atoms within the free residues from the current peptide chemical graph.
    \item ``UpdateTime'' and ``AlignTime'': As mentioned in \cref{subsec:sampling}, due to the sampling mechanism, updates of side-chain atoms within free residues might not be continuous. In other words, the coordinates of these atoms are occasionally updated, resulting in atoms having different times (or noise levels). Therefore, ``UpdateTime'' is introduced to store the individual time, and ``AlignTime'' is introduced to align the atom coordinates from different times to the same point. It's important to note that in ``AlignTime'', no model is involved as the previously cached denoised structures are reused.
\end{compactitem}

\renewcommand{\algorithmicrequire}{\textbf{Input:}}
\renewcommand{\algorithmicensure}{\textbf{Output:}}
\begin{algorithm}
\caption{Routed Sampling}
       \label{alg:routed_sampling}
        \begin{algorithmic}[1]
        \REQUIRE number of residues within the ligand peptide $N$, cyclization Type $O$,  3D receptor structures $\gT$, SDE solver time interval $\ud t$, 
            infinitesimal constant $\epsilon$
        \ENSURE cyclic peptide with its all-atom coordinates $\rvx_0$, chemical graph $\gG_0$, amino acid sequence $\gA_0$
\STATE $[\rmX_1,\rvx^O_1] \leftarrow \text{HarmonicPrior}(N,O,\gT)$ 
        \hfill $\rhd$ Initialize time-dependent atom73 state $\rmX_1$ and cyclization state $\rvx^O_1$
        
\STATE $\widetilde{\rvx}^O \leftarrow \text{Copy}(\rvx^O_1)$
        \hfill $\rhd$ Initialize denoised cyclization-related atom coordinates 
        $\widetilde{\rvx}^O$

\STATE $\gA_1 \leftarrow \text{Uniform}(20,N)$ 
        \hfill $\rhd$ Randomly initialize residue types not constrained by cyclization
\STATE $\gG_1 \leftarrow \text{Assemble}(\gA_1,O)$ 
        \hfill $\rhd$ Derive initial chemical graph 
\STATE $\rmT \leftarrow \mathbf{1}$  
        \hfill $\rhd$ Initialize a timer that records time for each atom in atom73 state
\STATE $t \leftarrow 1$ 
\WHILE{$t > \epsilon$}  
\STATE $\rvx_{t} \leftarrow \text{Extract}(\rmX_t,\gA_t)\, \cup\, \rvx^O_t $ 
            \hfill $\rhd$ Obtain all-atom $\rvx_t$ structure of current noisy peptide  
\STATE $\widehat{\rvx}_0 \leftarrow \text{\dockmodel}(\rvx_t, \gG_t, t)$ 
            \hfill $\rhd$ Predict denoised all-atom structure $\widehat{\rvx}_0$ structure  
            
\STATE $\widetilde{\rvx}^O \leftarrow \text{Cache}(\widetilde{\rvx}^O, \widehat{\rvx}_0, O)$ 
            \hfill

\STATE $\rvx_{t-\ud t} \leftarrow \text{Noise}(\widehat{\rvx}_0, \gG_t,  t\!-\!\ud t)$   
            \hfill $\rhd$ $\gG_t$ is required by harmonic noise

\STATE $\rmX_{t-\ud t} \leftarrow \text{Cache}(\rmX_t, \rvx_{t-\ud t}, \gA_t)$ 
            \hfill $\rhd$ Update $\rmX_t$ by saving new structures to specific states according to $\gA_t$

\STATE $\rvx^O_{t-\ud t} \leftarrow \text{Cache}(\rvx^O_{t}, \rvx_{t-\ud t}, O)$ 
            \hfill 
            
\STATE $\rmT \leftarrow \text{UpdateTimer}(\rmT, \gA_t, t\!-\!\ud t)$ 
            \hfill $\rhd$ Update the timer for the newly-updated atoms to the latest time
\STATE $\widetilde{\gG} \leftarrow \text{Subgraph}(\gG_t, \gA_t, O)$ 
            \hfill $\rhd$ Hide side chains of residues not constrained by cyclization 
\STATE $\gA_{t-\ud t} \leftarrow \text{\seqmodel}(\widehat{\rvx}_0, \widetilde{\gG}, t)$ 
            \hfill $\rhd$ Predict sequence based on the denoised structure $\widehat{\rvx}_0$

\STATE $\gG_{t-\ud t} \leftarrow \text{Assemble}(\gA_{t-\ud t}, O)$ 
            \hfill $\rhd$ Derive a new chemical graph given predicted sequence and cyclization

\STATE $\rvt \leftarrow \text{Extract}(\rmT, \gA_{t-\ud t})$ 
            \hfill $\rhd$ Obtain atom-wise time $\rvt$ (Atom might have different time)

\STATE $\rvx_{t- \ud t} \leftarrow \text{Extract}(\rmX_{t-\ud t}, \gA_{t-\ud t}) \, \cup \, \rvx^O_{t-\ud t}$ 
        \hfill $\rhd$ Align atoms with different time $\rvt$ to the same time $t\!-\!\ud t$
            
\STATE $\rvx_{t-\ud t} \leftarrow \text{AlignTime}(\rvx_{t-\ud t}, \widehat{\rvx}_0, \gG_{t-\ud t}, t\!-\!\ud t, \rvt)$ 
\hfill according to the reverse-time harmonic SDE 

\STATE $\rmX_{t-\ud t} \leftarrow \text{Cache}(\rmX_{t-\ud t}, \rvx_{t-\ud t}, \gA_{t-\ud t})$ 
            \hfill $\rhd$ Store the time-aligned atom coordinates
\STATE $t \leftarrow t - \ud t$
\ENDWHILE

\STATE $\rvx_{t} \leftarrow \text{Extract}(\rmX_t,\gA_t)\, \cup\, \rvx^O_t $ 
            
\STATE $\rvx_0 \leftarrow \text{\dockmodel}(\rvx_t, \gG_t, t)$   
    \hfill $\rhd$ Predict the all-atom structure finally  

\STATE $\gG_{0} \leftarrow \gG_t$ 

\STATE $\gA_{0} \leftarrow \gA_t$ 
        \end{algorithmic}
\end{algorithm}

\section{Proof of Prior Distribution Induced by Harmonic SDE}
\label{app:proof}
The difference between the widely-used SDE in SDE generative models and our introduced harmonic SDE is that the our perturbation process is anisotropic. Hence, here we provide the derivation of the prior distribution induced by the harmonic SDE in a similar proof by \citet{song2021scorebased}.

In \cref{eq:forward_harmonic_sde}, we define the $\gG_C$-dependent forward SDE as follows:
\begin{align}
    \ud\rvx^L = -\frac{1}{2}\beta(t)\rvx^L \, \ud t + \sqrt{\beta(t)}{\mLambda}^{\frac{1}{2}}\rmP^\intercal\,\ud \rvw, \nonumber
\end{align}
where $\beta(t)$ is a positive time-dependent scalar function,   $\rmP$ is an orthogonal matrix (i.e., $\rmP\rmP^\intercal=\rmI$), $\mLambda=\text{diag}(\lambda_1,\ldots,\lambda_{N_L})$ is a diagonal matrix that contains the eigenvalues, and $\rmH=\rmP\mLambda\rmP^\intercal$.

We denote the variance of the random variable $\rvx^L$ as  $\mathbf{\Sigma}(t)$, i.e., $\mathbf{\Sigma}(t) \coloneqq \operatorname{Cov}[\mathbf{x}(t)]$ for ${t\in[0,1]}$. The aforementioned SDE, characterized by affine drift and diffusion coefficients, allows us to employ Eq.~(5.51) from \citet{sarkka2019applied} to derive an ODE that describes the evolution of variance as follows:
\begin{align*}
    \frac{\mathrm{d} \mathbf{\Sigma}}{\mathrm{d} t} & = \beta(t) \big({({\mLambda}^{\frac{1}{2}}\rmP^\intercal)}^\intercal{\mLambda}^{\frac{1}{2}}\rmP^\intercal - \mathbf{\Sigma}(t)\big), \\
    & = \beta(t) (\rmH - \mathbf{\Sigma}(t)).
\end{align*}

Solving the above ODE, we derive 
\begin{align}
    \mathbf{\Sigma}(t) = \mathbf{H} + e^{\int_0^t - \beta(s) \mathrm{d} s}(\mathbf{\Sigma}(0) - \mathbf{H}), \nonumber
\end{align}
Once the boundary condition $\rvx^L_0$ is given, we have $\mathbf{\Sigma}(0)=\mathbf{0}$. Thus, the induced perturbation kernel has an analytic form as:
\begin{align}
    p_{0t}({\rvx}^L_t|{\rvx}^L_0) = \gN({\rvx}^L_t; 
    {\rvx}^L_0 
    e^{-\frac{1}{2}\int_0^t \beta(s) \ud s}, 
    \rmH - \rmH e^{-\int_0^t \beta(s) \ud s}
    ). \nonumber
\end{align}
Given $\lim_{t\to 1} \int_0^t \beta(s) \ud s =\infty$, the above perturbation process arrives at the prior distribution $p_1(\rvx^L_1) = \gN(\rvx^L_1;\mathbf{0},\rmH$).

\section{Implementation Details}
\label{app:implementation_details}

\subsection{Model Architecture}
\label{app:model_architecture}
Given a noisy sample at time $t$, two graphs are built for message passing with an SE(3)-equivariant neural network, which is parameterized by $\phi_K,\phi_B,\phi_C,\phi_H,\phi_E,\psi_K,\psi_C$ as introduced below. The $i$-th atom in the complex is attributed with an initial feature $\rvh_i$ and the bond~$ij$ in the noisy ligand is attributed with an initial feature $\rvb_{ij}$.
We first construct a k-nearest neighbor (knn) graph $\gG_K$ for the complex (i.e., the protein and the noisy ligand at time $t$), where each ligand atom is connected with the k-nearest atoms in the complex, to capture the protein-ligand interaction:
\begin{align}
\textstyle
    \Delta \rvh^K_i \leftarrow \sum_{j\in \gN_K(i)}  \phi_{K}(\rvh_i, \rvh_j, \Vert\rvx_i - \rvx_j \Vert, E_{ij},t),
    \nonumber
\end{align}
where $\gN_K(i)$ is the neighbors of atom $i$ in $\gG_K$, $E_{ij}$ indicates the edge $ij$ is a protein-protein, ligand-ligand or protein-ligand edge.

We also leverage the chemical graph $\gG_C$ of the ligand as we have defined previously to make the model aware of the connection information introduced by the chemical bonds: 
\begin{align}
    \rve_{ij} \leftarrow \phi_{B} (\Vert \rvx_i - \rvx_j, \rvb_{ij}\Vert),
    \nonumber
    \\
\textstyle
    \rvh^C_{i} \leftarrow \sum_{j\in\gN_C(i) \phi_{C}}(\rvh_i,\rvh_j,\rve_{ij},t).
    \nonumber
\end{align}
We further aggregate the hidden features of ligand atoms and bonds from these two graphs as follows:
\begin{align}
    \rvh_i \leftarrow \rvh_i + \phi_H(\Delta\rvh^K_i + \Delta\rvh^C_i),
    \nonumber
    \\
    \textstyle
    \rvb_{ij} \leftarrow \sum_{k\in \gN_C(j)\backslash \{i\}} \phi_B(\rvh_i,\rvh_j, \rvh_k, \rve_{ik}, \rve_{kj},t).
    \nonumber
\end{align}
Finally, we update the ligand atom positions as follows:
\begin{align}
\textstyle
    \Delta \rvx^K_{i} \leftarrow \sum_{j\in\gN_K(i)}(\rvx_j-\rvx_i)\psi_K(\rvh_i,\rvh_j,\Vert\rvx_i-\rvx_j\Vert,t),
    \nonumber
    \\
\textstyle
    \Delta \rvx^C_{i} \leftarrow \sum_{j\in\gN_C(i)}(\rvx_j-\rvx_i)\psi_K(\rvh_i,\rvh_j,\Vert\rvx_i-\rvx_j\Vert,\rve_{ij},t),
    \nonumber
    \\
    \rvx_i \leftarrow \rvx_i + (\Delta\rvx_i^K + \Delta \rvx_i^C) \cdot \mathds{1}\{i\in \gG_C\},
    \nonumber
\end{align}
where $\mathds{1}\{i\in \gG_C\}$ indicates whether atom $i$ belongs to the ligand since the protein atom positions are fixed and we only update ligand atom positions.

We denote the final output of the SE(3)-equivariant neural network as $\mD_\vtheta(\rvx^L_t,t)$, where $\mD_\vtheta$ is composed of $\phi_K,\phi_B,\phi_C,\phi_H,\phi_E,\psi_K,\psi_C$ as introduced above. 

\subsection{Training Details}
We use the same optimizer setting for both \dockmodel and \seqmodel: AdamW \citep{loshchilov2017decoupled} optimizer with constant learning rate 0.0001, beta1 0.9, beta2 0.999, and weight decay 0.01. 
For beta schedule, we use $\beta(t)=(\beta_{\text{max}}-\beta_{\text{min}}) t + \beta_{\text{min}}$, where $\beta_\text{min}=0.01$ and $\beta_\text{max}=3.0$. Note that $\lim_{t\to 1}\int_0^s \beta(s) \ud s$ is sufficiently large compared to the variance of our data distribution. To train \dockmodel, we sample $t\sim \gU[0,1]$. To train \seqmodel, we sample $t\sim \gU[0,0.5]$ due to the fact that the denoised structure output by trained model $\dockmodel$ at time $t=0.5$ or more has extremely limited information to determine the residue types. \dockmodel converges within 48 hours and \seqmodel converges within 18 hours on 8 NVIDIA H100 GPUs.

\subsection{Sampling Details}
For routed sampling, we divide time interval  $[0,1]$ into 1,000 steps. Inspired by \citet{chu2024all}, we skip \seqmodel when $t>0.5$ since the structures are too noisy to provide sufficient information for residue type prediction. This approach also accelerates the generative process and reduces the computational cost of inference. When $t<0.5$, at each step, sequence and structures are iteratively updated by \seqmodel and \dockmodel, respectively.

\section{Experimental Details} 
\label{app:experimental_details}

\subsection{Relaxation and Energy Estimation} 
Cyclic peptides offer notable advantages in terms of both system stability and binding affinity. In specific, the stability of a protein-peptide complex is inversely proportional to its overall free energy, with lower free energy indicating greater stability. To assess this, the \texttt{FastRelax} protocol in PyRosetta~\citep{chaudhury2010pyrosetta} is employed to relax each complex, after which the total energy is evaluated using the REF2015 scoring function. Binding affinity is measured with the \texttt{InterfaceAnalyzerMover}, which calculates the binding energy at the interface between the peptide and the target protein within the relaxed complex. An increase in binding energy reflects enhanced peptide binding affinity, suggesting potential functional improvements.

For each target, linear peptide methods generate 8 samples with the golden peptide length (i.e., the number of residues within reference linear peptide). Cyclic peptide methods, lacking a reference length, enumerate residues from 5 to 20 (or 8 to 23 for side-to-side cyclic peptides), generating 2 samples per length. All reference ligands and samples are relaxed and evaluated using a standard scoring method as described above. For each target and method, we apply the Borda method to select the best ligand, accounting for both stability and affinity, which is then reported in the final results.

\subsection{Detailed Experimental Results}
We have provided detailed energy measurement results for each target, including the reference ligand, linear peptides designed by baselines, and cyclic peptides designed by our methods, in \cref{tab:appendix_evaluation_all_1,tab:appendix_evaluation_all_part_2}.

\subsection{Inference Speed}

We benchmark the average time of generating one peptide for all co-design baselines and our methods on a single NVIDIA A100-SXM4-80GB GPU. The results are reported in \cref{tab:generation_time}. Given that computational drug design does not demand real-time model response, the inference time of our method is deemed acceptable.

\definecolor{darkgreen}{RGB}{34,139,34}
\begin{table}[h]
    \centering
    \caption{Generation time of all co-design baselines and our methods.}
    \begin{adjustbox}{width=0.38\textwidth}
    \renewcommand{\arraystretch}{1.2}
    \begin{tabular}{l|c|c}
    \toprule
    Method & Peptide Type & Time (s) \\
    \midrule
    
    ProteinGenerator  & Linear & 31.80 \\
    PepFlow & Linear &12.09 \\
    PepGLAD  & Linear& 4.40 \\
    \method  & Cyclic& 16.88 \\
    \bottomrule
    \end{tabular}\label{tab:generation_time}
    \renewcommand{\arraystretch}{1}
    \end{adjustbox}
\end{table}

\subsection{Evaluation on Linear Peptide Design}
While designing linear peptides is not our primary focus, we compare our method with existing baselines in this task. 

Unlike the variable-length setting for cyclic peptides, the task of linear peptide design can leverage known reference peptide lengths for target proteins in the test set. For fair comparison across methods, we sample 8 linear peptides per target matching the reference length and relax the complex structure by Rosetta \citep{chaudhury2010pyrosetta,alford2017rosetta}. We then apply the Borda method to choose the optimal linear peptides, considering both stability and affinity. 
We report the average and median for Stability and Affinity of the linear peptides for targets in the test set. We also report the average for the fraction of hydrophobic and charged residues (relevant for specificity) \citep{ye2024proteinbench}, DockQ, iRMSD, LRMSD, BSR, and Diversity.
We use DockQ package~\footnote{\url{https://github.com/bjornwallner/DockQ}} to compute DockQ, iRMSD, and LRMSD. We follow the definition of binding site ratio (BSR) in \citet{li2024full}. A lower hydrophobic/charged ratio indicates a lower risk of non-specific binding \citep{makowski2024optimization}. The results are reported in \cref{tab:linear_peptide}. Notably, the fraction of hydrophobic and charged residues of our designed peptides resembles that of reference. Our method also shows superiority in structural properties.

\definecolor{darkgreen}{RGB}{34,139,34}
\begin{table}[h]
    \centering
    \caption{Summary of properties of reference peptides, linear peptides designed by baseline methods and \method. ($\downarrow$) / ($\uparrow$) denotes a smaller / larger number is better.}
    \begin{adjustbox}{width=1\textwidth}
    \renewcommand{\arraystretch}{1.2}
\begin{tabular}{l|c|l|cc|cc|c|c|c|c|c|c|c   }
    \toprule
    \multirow{2}{*}{Method}
    & \multirow{2}{*}{Co-Design}
    & \multirow{2}{*}{Peptide Type} 
    & \multicolumn{2}{c|}{Stability ($\downarrow$)} & \multicolumn{2}{c|}{Affinity ($\downarrow$)} 
    & Hydrophobic  
    &	Charged 
    &	DockQ
    &	iRMSD
    &	LRMSD	
    &   BSR
    & Diversity \\
    & & & Avg. & Med. & Avg. & Med. & Ratio ($\downarrow$) & Ratio ($\downarrow$) & ($\uparrow$) & ($\downarrow$) & ($\downarrow$) & ($\uparrow$) & ($\uparrow$) \\
    \midrule
     {Reference} & N/A &  Linear & -672.53 & -634.71 & -85.03 & -78.70 & 0.48	&0.28 & N/A & N/A & N/A & N/A & N/A \\ 
    \midrule
    RFDiffusion & \XSolidBrush  & Linear  & -633.51 & -607.82 & -70.30 & -61.35 & 0.59	&0.27	&0.18	&5.37	&20.10	&0.33 & 0.55\\
    ProteinGenerator & \Checkmark & Linear & -576.39 & -554.70 & -46.98 & -40.39  &  0.53	&0.32	&0.12	&5.56	&23.97	&0.20 & 0.58\\
    PepFlow & \Checkmark & Linear   &  -576.16 & -498.31 & -47.88 & -42.40 &  0.60	&0.17	&0.44	&2.49	&9.42	&0.56 & 0.70\\
    PepGLAD & \Checkmark & Linear   & -359.44  & -310.33 &  -45.06 &  -38.56 &  0.53	&0.25	&0.30	&2.68	&11.99	&0.39 &0.79 \\
    \method  & \Checkmark & Linear & -567.34 &  -510.58	 & -55.48 & -49.89 &  0.45	&0.24	&0.32	&2.36	&9.91	&0.60 & 0.77\\
    \bottomrule
    \end{tabular}
\label{tab:linear_peptide}
    \renewcommand{\arraystretch}{1}
    \end{adjustbox}
\end{table}

\subsection{Ablation Studies} 
\label{app:ablation_studies}

\textbf{Effects of \seqmodel .} We study the effects of \seqmodel compared with the following two setups: ``w/ fix seq'' where the residue types are randomly sampled and fixed with only atom coordinates updated during the generative process, ``w/ random seq'' where the residue types are randomly sampled from a uniform distribution instead of predicted by \seqmodel during the generative process. The results are shown in \cref{tab:ablation_seqmodel}. It can be observed that both variants perform worse than \method, which demonstrates that \seqmodel can effectively discover critical residue types for protein-ligand interaction. ``w/ random seq'' performs the worst, possibly because random residue types offer no information gain, and the residue types are updated too frequently. This frequent updating hinders the \dockmodel from effectively updating the side-chain atoms of the free residues.

\definecolor{darkgreen}{RGB}{34,139,34}
\begin{table}[h]
    \centering
    \caption{Ablation study on the effect of \seqmodel.}
    \begin{adjustbox}{width=0.44\textwidth}
    \renewcommand{\arraystretch}{1.2}
    \begin{tabular}{l|cc|cc}
    \toprule
    \multirow{2}{*}{Method} 
    & \multicolumn{2}{c|}{Stability ($\downarrow$)} & \multicolumn{2}{c}{Affinity ($\downarrow$)} \\
    & Avg. & Med. & Avg. & Med.   \\
    \midrule
    \method & -568.04 &  -519.66 & -50.86 & -46.62  \\
    w/ fix seq & -525.73 &  -439.69 & -41.56 & -38.75  \\
    w/ random seq & -521.48 &  -425.17 & -38.64 & -39.44  \\
    
    \bottomrule
    \end{tabular}\label{tab:ablation_seqmodel}
    \renewcommand{\arraystretch}{1}
    \end{adjustbox}
\end{table}

\textbf{Effects of Harmonic SDE.} We study the effects of harmonic SDE. We introduce a variant with isotropic Gaussian as prior and noise distribution (denoted as ``w/o Harmonic''). The results are shown in \cref{tab:ablation_seqmodel} and validate the effectiveness of harmonic prior and noise. 
To explore the underlying reasons, we examined the trajectories of the generative process and found that the harmonic prior provides a good initialization for atom positions, where bonded atoms are located nearby. This feature might be beneficial for routed sampling because some side-chain atom updates can be discontinuous, and such correlated initialization helps mitigate the errors induced by these discontinuous updates.

\definecolor{darkgreen}{RGB}{34,139,34}
\begin{table}[h]
    \centering
    \caption{Ablation study on the effect of Harmonic SDE.}
    \begin{adjustbox}{width=0.44\textwidth}
    \renewcommand{\arraystretch}{1.2}
    \begin{tabular}{l|cc|cc}
    \toprule
    \multirow{2}{*}{Method} 
    & \multicolumn{2}{c|}{Stability ($\downarrow$)} & \multicolumn{2}{c}{Affinity ($\downarrow$)}  \\
    & Avg. & Med. & Avg. & Med.   \\
    \midrule
    \method & -568.04 &  -519.66 & -50.86 & -46.62  \\
    w/o Harmonic & -534.38 & -439.23 & -39.43 & -41.08 \\
    \bottomrule
    \end{tabular}\label{tab:ablation_harmonic_sde}
    \renewcommand{\arraystretch}{1}
    \end{adjustbox}
\end{table}

\subsection{System Setup and Protocols of Molecular Dynamics Simulation}
\label{app:md_setup}
To simulate the protein-peptide systems, hydrogen atoms are added, and the dominant protonation state of titratable residues at pH 7 is determined using PropKa in PDB2PQR~\citep{dolinsky2007pdb2pqr}. Subsequently, the systems are solvated in a 10 \text{\AA} truncated water box, with sodium and chloride ions added to neutralize the system at a concentration of 150 mM to mimic physiological saline. The ff14SB ~\citep{maier2015ff14sb} parameter set is applied to proteins and peptides, and the TIP3P model is used for water~\citep{jorgensen1983comparison, li2024delineating}.

All simulations were run on RTX 4090 GPUs using the CUDA implementation of particle-mesh Ewald (PME) molecular dynamics in Amber22~\citep{salomon2013routine}. At first, to relax each system thoroughly, two stages of energy minimization are performed. In the first stage, 2,500 steepest descent and 2,500 conjugate gradient cycles were applied to all atoms, with constraints on water molecules and counterions. In the second stage, the same cycles were repeated without constraints. Initial velocities are randomly sampled from a Boltzmann distribution. The systems are then heated from 0 K to 310 K over 500 ps in the NVT ensemble, using a Langevin thermostat and harmonic restraints of 10.0 $\text{kcal}\cdot\text{mol}^{-1}\cdot\text{\AA}^{-2}$. During equilibration at 300 K and 1 bar under NPT conditions, harmonic restraints on protein and peptide atoms were progressively reduced from 5.0 to 0.1 $\text{kcal}\cdot\text{mol}^{-1}\cdot\text{\AA}^{-2}$ in four steps at 0.5 ns intervals, totaling 2.5 ns. All restraints are completely removed during production simulation under 310K and 1 bar, which are maintained using the Langevin thermostat and Berendsen barostat, respectively. A timestep of 4.0 fs is used with hydrogen mass repartitioning~\citep{hopkins2015long}. Bond lengths are constrained via SHAKE~\citep{ryckaert1977numerical}, and non-bonded interactions are cut off at 10~\text{\AA}.

\subsection{Visualization of Structure Ensembles Simulated by Molecular Dynamics}
\begin{figure}[h]
    \centering
    \includegraphics[width=0.84\linewidth]{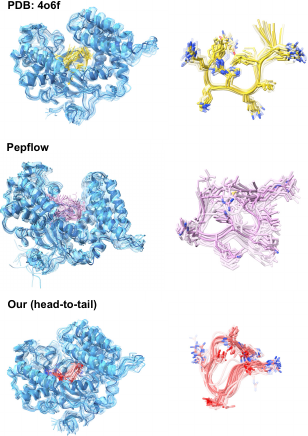}
    \caption{Structure ensembles of SMYD2 from multiple perspectives.}
    \label{fig:md_different_4o6f}
\end{figure}

\begin{figure}[h]
    \centering
    \includegraphics[width=0.84\linewidth]{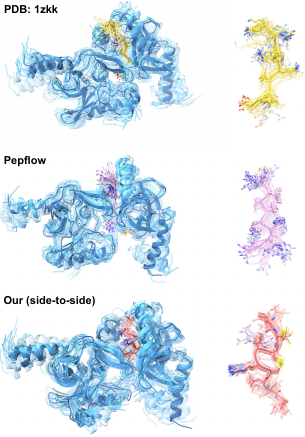}
    \caption{Structure ensembles of SET8 from multiple perspectives.}
    \label{fig:md_different_1zkk}
\end{figure}

We present additional views of the structure ensembles generated by molecular dynamics simulations in \cref{fig:md_different_4o6f} and \cref{fig:md_different_1zkk}.

\section{Examples of Designed Cyclic Peptides}

\begin{figure}[h]
\begin{center}
\centerline{\includegraphics[width=0.98\textwidth]{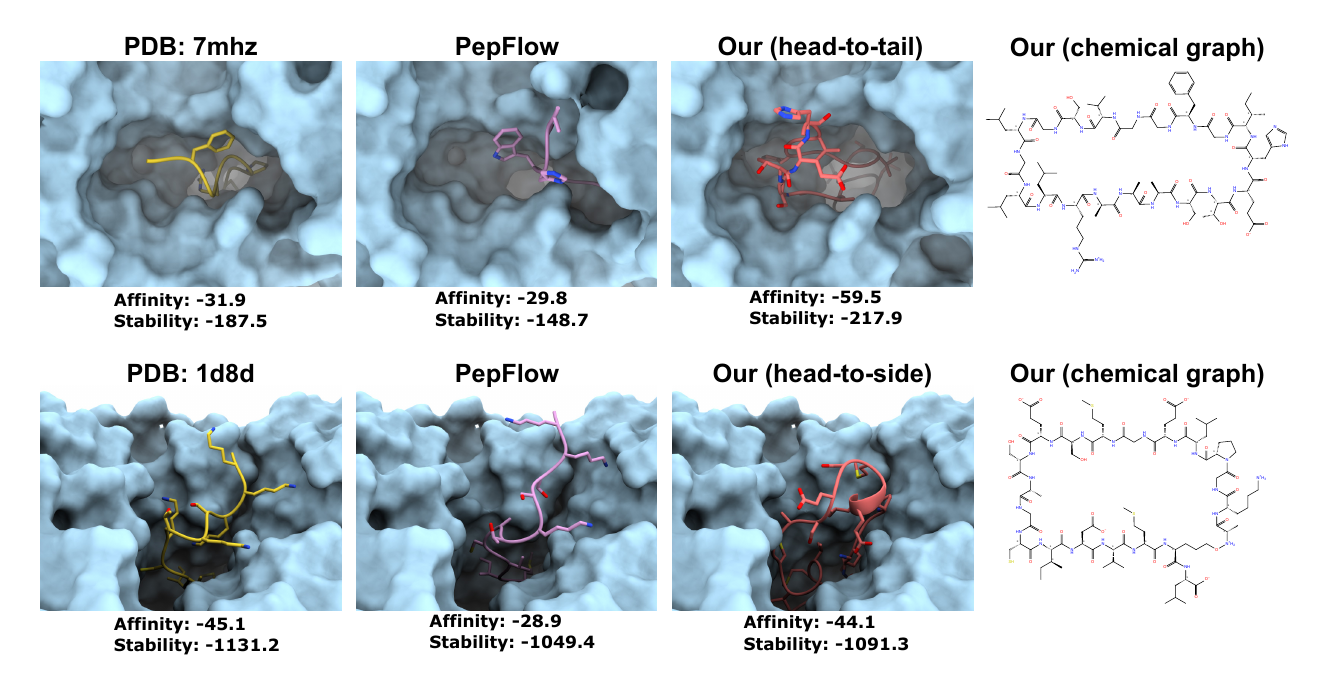}}
\caption{Visualization of reference ligand, linear peptides designed by PepFlow, and cyclic peptides designed by \method.}
\label{fig:main_gen_example}
\end{center}
\end{figure}

\begin{figure}[h]
    \centering
    \includegraphics[width=0.94\linewidth]{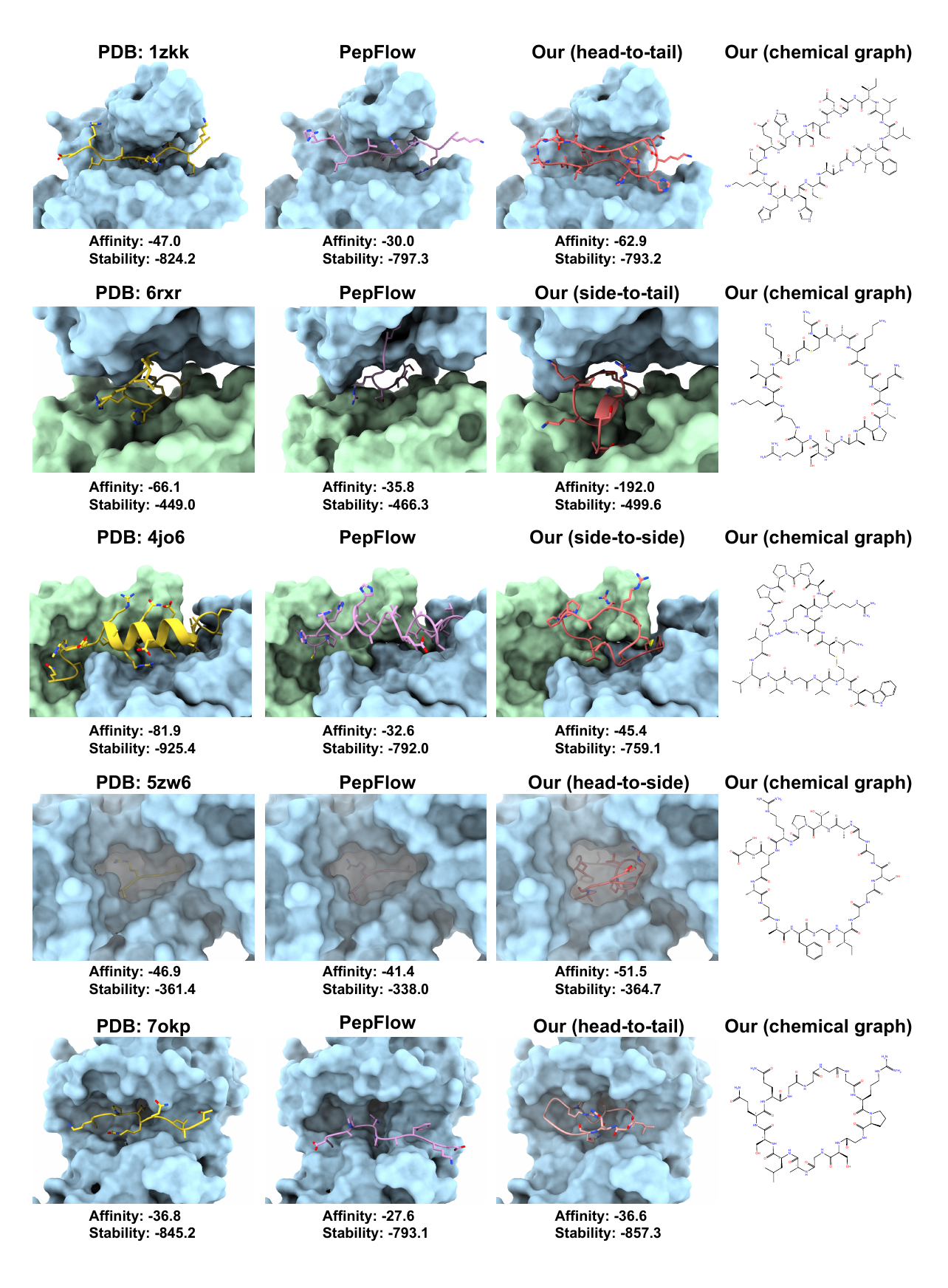}
    \caption{Examples of reference peptides, linear peptides designed by PepFlow, and cyclic peptides designed by our method.}
    \label{fig:generated_example_1}
\end{figure}

\begin{figure}[h]
    \centering
    \includegraphics[width=0.94\linewidth]{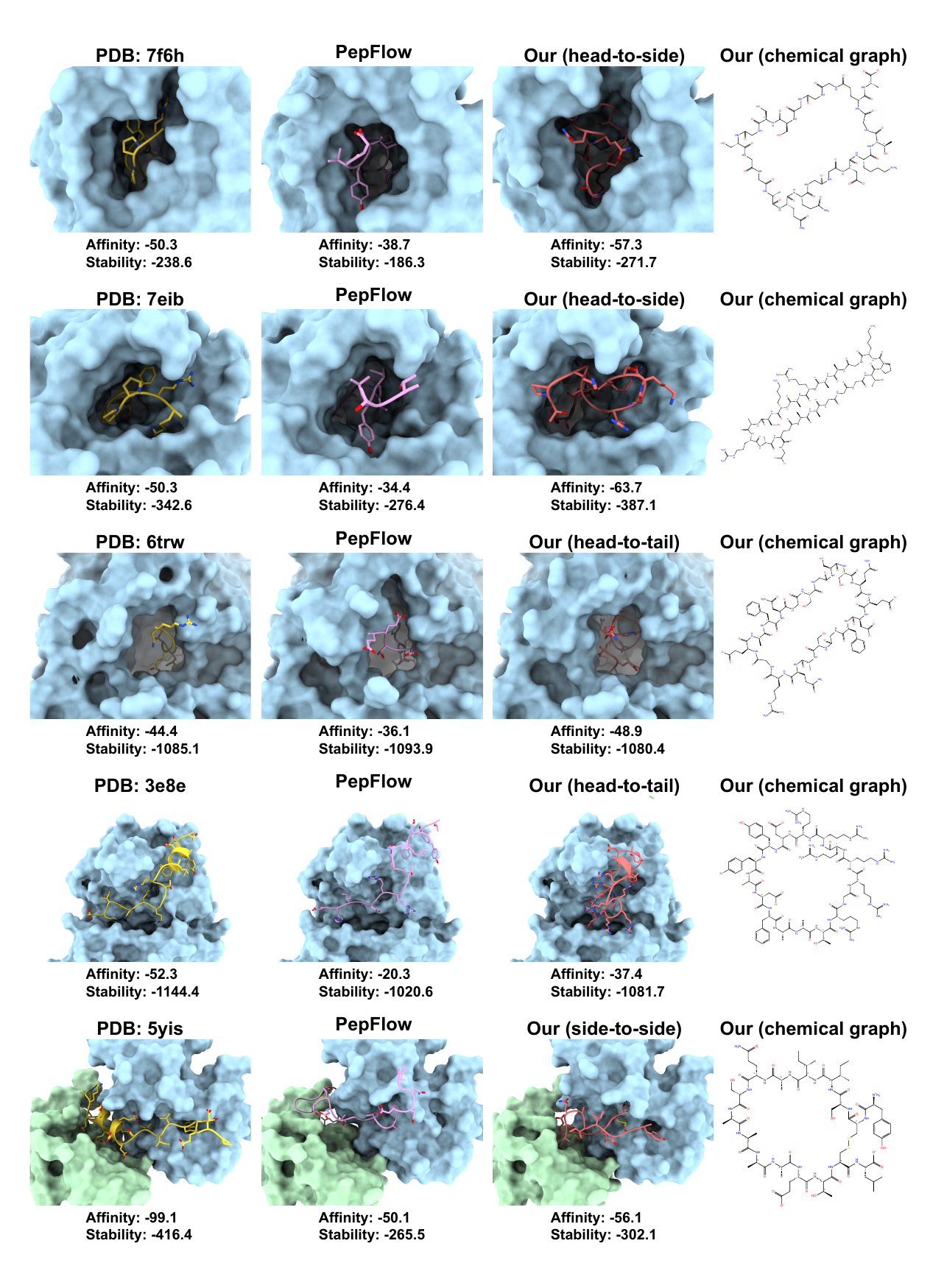}
    \caption{Examples of reference peptides, linear peptides designed by PepFlow, and cyclic peptides designed by our method.}
    \label{fig:generated_example_2}
\end{figure}

Here, we present more compelling results of our generated cyclic peptides targeting different receptors in \cref{fig:main_gen_example,fig:generated_example_1,fig:generated_example_2}. We find that our generated cyclic peptides consistently exhibit higher or competitive affinities with greater interaction stabilities when binding to the receptor. In contrast, linear peptides sampled from PepFlow often result in unstable structures and weaker binding. Furthermore, our designed cyclic peptides not only interact with key receptor regions, similar to linear peptides and native peptides, but also establish new, stable, and tight interactions in additional regions. Additionally, our generated 3D cyclic peptides consistently align well with the corresponding 2D chemical graphs, highlighting the effective integration of our two models.

\begin{figure}[h]
    \centering
    \includegraphics[width=0.98\linewidth]{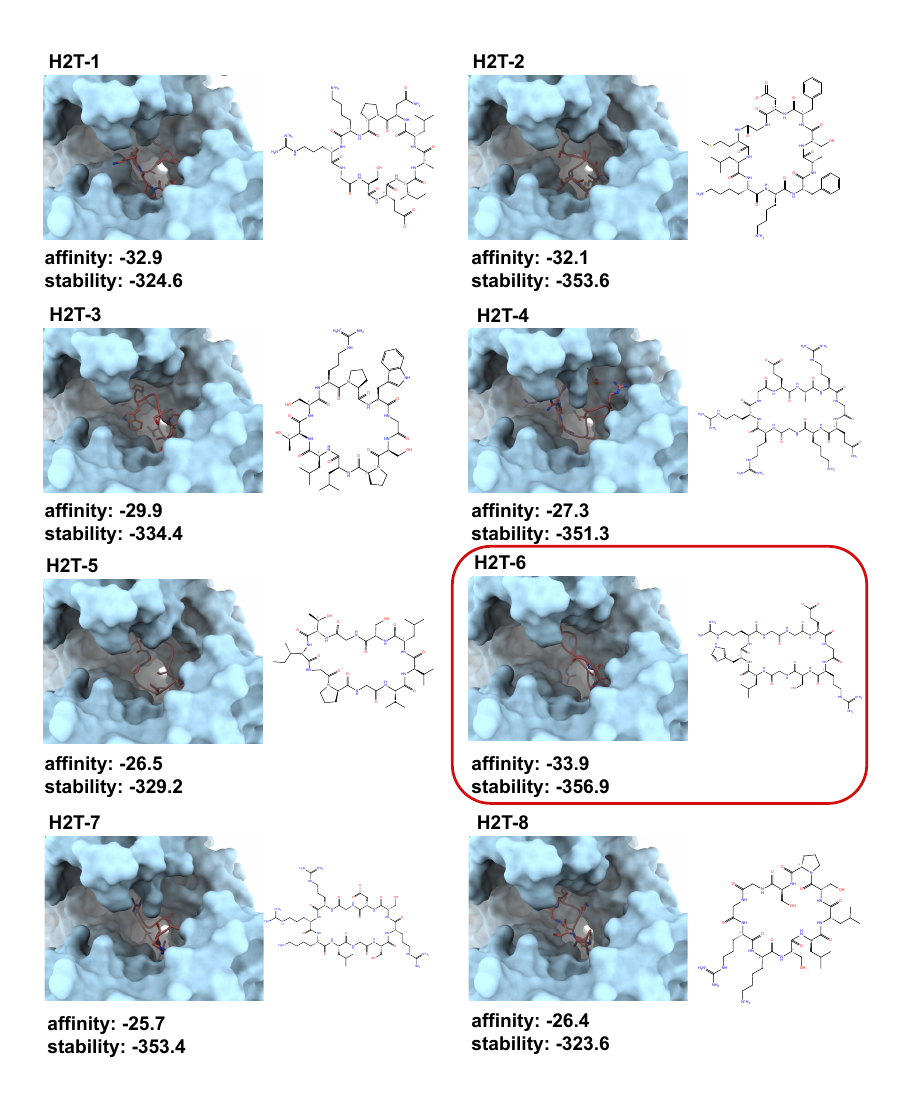}
    \caption{Head-to-tail cyclic peptides designed for target SMYD2.}
    \label{fig:4o6f_select}
\end{figure}

\begin{figure}[h]
    \centering
    \includegraphics[width=0.98\linewidth]{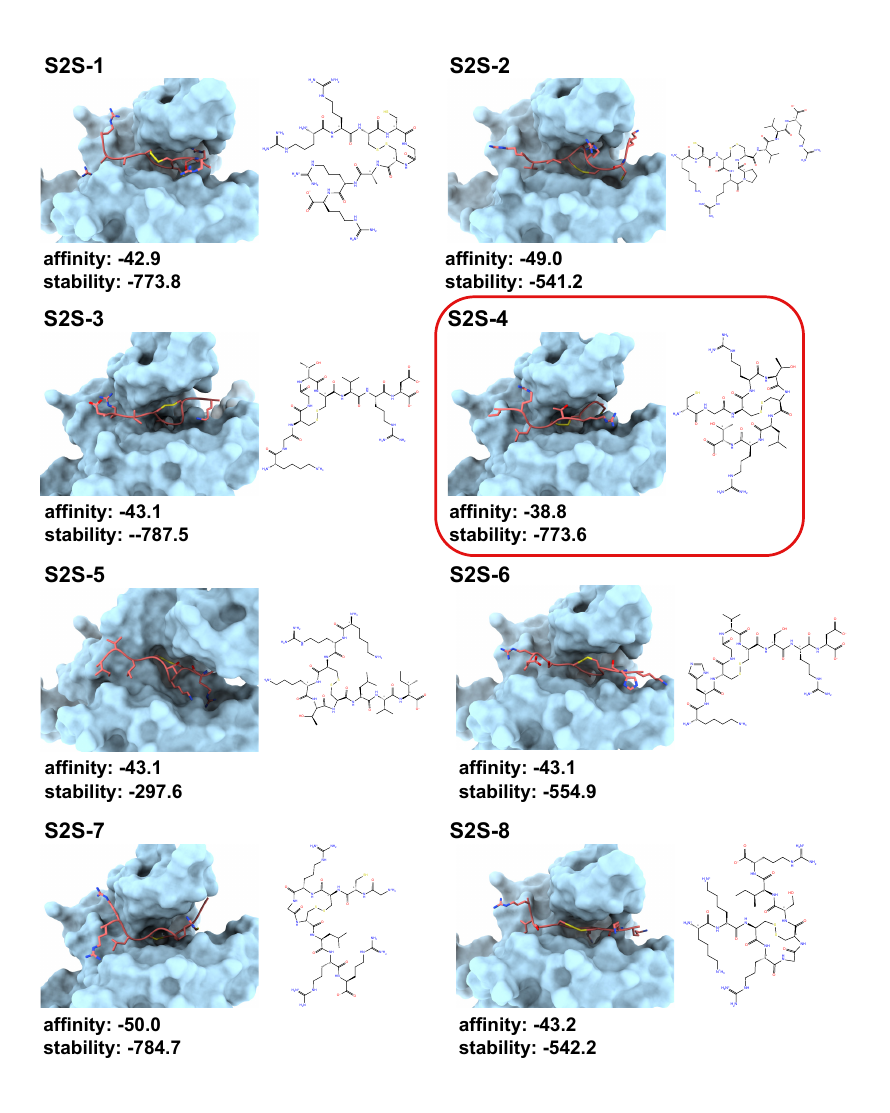}
    \caption{Side-to-side cyclic peptides designed for target SET8.}
    \label{fig:1zkk_select}
\end{figure}

\definecolor{darkgreen}{RGB}{34,139,34}
\begin{table*}[t!]
    \scriptsize
    \centering
    \caption{Stability and affinity of the reference peptide, linear peptides designed by baseline methods, and cyclic peptide designed by our method along with the cyclization type.}
    \renewcommand{\arraystretch}{1.2}
    \begin{tabular}{l|rr|rr|rr|rr|rr|rrr}
    \toprule
    \multirow{2}{*}{Target}
    & \multicolumn{2}{c|}{Reference}
    & \multicolumn{2}{c|}{RFDiffusion}
    & \multicolumn{2}{c|}{ProteinGenerator}
    & \multicolumn{2}{c|}{PepFlow}
    & \multicolumn{2}{c|}{PepGLAD}
    & \multicolumn{3}{c}{Our}
    \\
    \cmidrule(lr){2-3} \cmidrule(lr){4-5} \cmidrule(lr){6-7} 
    \cmidrule(lr){8-9} \cmidrule(lr){10-11} \cmidrule(lr){12-14}
    & Stab. & Affi. & Stab. & Affi. & Stab. & Affi. & Stab. & Affi. & Stab. & Affi. & Stab. & Affi. & Type \\
    \midrule
1d8d & -1131.2 & -45.1 & -1092.2 & -37.2 & -975.5 & -43.7 & -1049.4 & -28.9 & -1070.0 & -30.1 & -1091.3 & -44.1 & h2s \\
1hr8 & -417.7 & -48.4 & -363.9 & -78.7 & -370.0 & -39.3 & -428.8 & -36.8 & -207.8 & -28.5 & -313.1 & -37.2 & s2t \\
1rgq & -374.0 & -140.2 & -390.1 & -153.6 & -289.1 & -121.1 & -167.5 & -69.3 & 82.6 & -67.2 & -255.1 & -83.5 & s2t \\
1vzj & -80.1 & -138.7 & 282.8 & -66.3 & 356.1 & -52.8 & 12.4 & -112.2 & 20.4 & -90.6 & 143.4 & -105.3 & h2s \\
1xoc & -899.3 & -66.9 & -763.8 & -40.7 & -866.0 & -41.2 & -886.2 & -51.1 & -778.8 & -30.8 & -867.3 & -47.7 & h2s \\
1zkk & -824.2 & -47.0 & -843.0 & -59.0 & -824.0 & -40.2 & -797.3 & -30.0 & -439.0 & -35.5 & -793.2 & -62.9 & h2t \\
2arq & -556.3 & -148.5 & -588.7 & -165.9 & -428.5 & -29.2 & -428.2 & -63.5 & -204.1 & -53.1 & -438.3 & -91.5 & h2s \\
2mpz & -1136.2 & -223.1 & -1114.8 & -203.7 & -1031.2 & -91.4 & -937.5 & -84.0 & 491.4 & -79.3 & -1059.0 & -244.1 & h2s \\
2vda & 223.6 & -59.6 & 336.7 & -33.5 & 393.3 & -32.4 & 343.5 & -35.1 & 960.0 & -47.8 & 373.1 & -25.9 & h2t \\
2wqj & -576.8 & -123.3 & -571.3 & -125.7 & N/A & N/A & -447.3 & -71.8 & 126.2 & -76.0 & -386.2 & -48.8 & h2t \\
2xjz & 95.0 & -164.0 & 234.9 & -102.2 & 343.4 & -58.1 & 251.8 & -72.0 & N/A & N/A & 375.2 & -46.6 & h2t \\
3e8e & -1144.4 & -53.3 & -1123.9 & -50.8 & -1119.1 & -39.1 & -1020.6 & -20.3 & -551.5 & -27.0 & -1081.7 & -37.4 & h2t \\
3ech & -569.4 & -101.1 & -541.2 & -78.8 & -476.6 & -84.8 & -446.8 & -44.0 & -231.2 & -65.0 & -436.8 & -56.0 & h2t \\
3ewf & -1453.3 & -36.4 & -1425.1 & -16.9 & -1433.9 & -11.7 & -1356.4 & 0.4 & -1390.7 & -11.9 & -1484.1 & -48.0 & h2s \\
3fii & -290.1 & -105.2 & -197.1 & -88.8 & -173.3 & -34.7 & -29.4 & -30.5 & -39.3 & -100.2 & -100.3 & -45.9 & h2t \\
3h8a & -528.3 & -71.1 & -538.8 & -68.6 & -446.2 & -57.7 & -400.1 & -58.1 & 232.1 & -38.3 & -442.8 & -45.2 & h2s \\
3j89 & -919.6 & -118.1 & -908.3 & -115.3 & -876.1 & -84.7 & -771.6 & -59.9 & N/A & N/A & -744.2 & -56.5 & h2t \\
3lk4 & -263.2 & -101.4 & -215.7 & -80.6 & -176.2 & -26.4 & -256.9 & -68.3 & N/A & N/A & -223.3 & -31.5 & h2s \\
3mhp & -1021.0 & -93.4 & -948.3 & -64.0 & -851.9 & -13.8 & -903.4 & -43.8 & -682.8 & -59.5 & -849.0 & -38.0 & h2t \\
3o0e & -353.1 & -43.0 & -286.4 & -55.7 & -338.0 & -42.9 & -357.9 & -33.1 & -126.2 & -30.8 & -332.9 & -30.4 & s2t \\
3pl7 & -215.7 & -105.9 & -266.8 & -124.2 & -257.3 & -104.2 & -140.4 & -98.1 & 398.6 & -98.9 & -81.0 & -59.1 & h2s \\
3ro2 & -606.5 & -86.2 & -376.7 & -32.1 & -345.0 & -31.6 & -539.1 & -52.5 & -117.2 & -45.8 & -378.1 & -42.9 & h2s \\
3ryb & -990.0 & -54.9 & -993.3 & -59.4 & -894.1 & -34.5 & -1003.8 & -48.2 & -857.7 & -37.0 & -942.6 & -47.4 & h2t \\
3twt & -965.3 & -57.3 & -862.1 & -23.2 & -851.4 & -14.4 & -936.2 & -31.5 & -782.6 & -31.8 & -892.3 & -36.6 & h2s \\
3vvs & -691.1 & -55.1 & -632.5 & -40.1 & -634.7 & -49.6 & -692.0 & -50.3 & -632.0 & -38.1 & -722.5 & -63.7 & h2s \\
3wy9 & -618.2 & -59.8 & -522.1 & -73.3 & -610.9 & -56.6 & -523.7 & -34.0 & -313.7 & -49.7 & -550.8 & -24.4 & h2t \\
3zha & -404.5 & -122.1 & -187.8 & -37.2 & -136.9 & -62.0 & -308.1 & -60.1 & -451.7 & -68.2 & -261.4 & -55.9 & h2s \\
4chg & -936.0 & -112.9 & -790.8 & -59.4 & -809.6 & -77.7 & -790.1 & -66.7 & N/A & N/A & -714.0 & -64.2 & h2t \\
4e7v & -490.4 & -125.4 & N/A & N/A & N/A & N/A & -308.0 & -73.0 & 53.7 & -96.0 & -473.1 & -158.2 & h2t \\
4edn & -800.6 & -75.1 & -815.0 & -75.6 & -694.9 & -24.8 & -760.6 & -44.3 & -521.4 & -35.6 & -711.4 & -53.3 & h2t \\
4hom & -1143.1 & -33.5 & -1108.8 & -56.5 & -1190.9 & -22.5 & -1108.6 & -36.9 & -1105.9 & -21.9 & -1184.8 & -34.2 & h2t \\
4jo6 & -925.4 & -81.9 & -927.8 & -72.0 & -861.7 & -40.6 & -792.0 & -32.6 & -131.5 & -60.4 & -759.1 & -45.4 & s2s \\
4m1c & -156.4 & -41.1 & -98.9 & -26.8 & -16.6 & -28.4 & -100.1 & -33.6 & -74.8 & -25.5 & -102.3 & -30.9 & h2s \\
4o6f & -439.3 & -49.9 & -422.1 & -66.9 & -425.3 & -49.2 & -428.8 & -32.3 & -370.9 & -21.7 & -425.6 & -44.1 & h2t \\
4po7 & -301.2 & -42.9 & -294.3 & -65.0 & -244.1 & -56.2 & -294.1 & -25.4 & -267.9 & -20.9 & -261.9 & -24.9 & h2t \\
4qae & -912.4 & -94.9 & -945.3 & -67.4 & -927.2 & -46.4 & -870.7 & -56.3 & -198.1 & -55.6 & -851.1 & -54.3 & s2s \\
4uqz & -733.4 & -96.8 & -652.8 & -30.9 & -640.8 & -35.5 & -557.0 & -39.6 & -555.3 & -38.1 & -628.2 & -48.0 & s2s \\
4wsi & -281.4 & -93.5 & -145.6 & -58.1 & -107.4 & -34.6 & -180.7 & -30.9 & -85.3 & -45.6 & -210.8 & -32.0 & h2t \\
4x3o & -495.0 & -28.0 & -494.6 & -30.6 & -420.4 & -21.3 & -441.6 & 0.0 & -468.4 & -16.8 & -585.4 & -83.4 & h2t \\
4xpd & -114.1 & -23.3 & -62.1 & -20.3 & -78.3 & -30.0 & -27.9 & -6.3 & 0.3 & -15.6 & -175.7 & -48.9 & h2t \\
4xtr & -660.2 & -110.7 & -659.0 & -97.4 & -662.6 & -77.0 & -594.0 & -81.9 & -372.5 & -57.4 & -534.9 & -62.6 & h2s \\
4yjl & -1013.7 & -73.3 & -859.8 & -39.6 & -814.8 & -24.0 & -934.8 & -38.7 & -745.0 & -31.4 & -923.3 & -30.8 & h2t \\
4zp3 & -384.2 & -102.1 & -414.8 & -119.2 & -228.4 & -42.2 & -228.8 & -61.0 & 24.9 & -71.6 & -214.1 & -39.5 & h2s \\
5apk & -367.8 & -62.1 & -240.0 & -65.1 & -146.3 & -62.7 & -310.0 & -48.0 & -269.8 & -33.2 & -330.1 & -60.7 & h2t \\
5brm & -469.6 & -72.2 & -379.6 & -38.4 & -375.3 & -24.4 & -405.0 & -42.6 & -301.6 & -44.9 & -414.8 & -43.8 & h2t \\
5c6h & -358.6 & -83.3 & -350.2 & -87.3 & -220.6 & -36.3 & -301.3 & -92.8 & -29.3 & -65.9 & -231.5 & -46.6 & h2t \\
5dhm & -589.5 & -155.4 & -607.8 & -149.4 & -451.7 & -64.2 & -483.1 & -39.4 & -123.1 & -77.9 & -424.9 & -53.1 & h2t \\
5e2q & -977.0 & -65.2 & -865.6 & -37.4 & -839.6 & -28.8 & -932.5 & -47.5 & -819.0 & -31.9 & -902.4 & -58.5 & s2s \\
5et1 & -1436.7 & -77.3 & -1321.0 & -41.4 & -1391.5 & -54.9 & -1318.7 & -32.9 & -592.4 & -37.1 & -1313.0 & -26.5 & h2t \\
5iyx & -744.4 & -59.5 & -612.4 & -7.7 & -652.9 & -37.8 & -659.4 & -34.7 & -564.2 & -28.3 & -653.3 & -38.1 & h2t \\
5j3t & -718.3 & -102.9 & -624.8 & -57.4 & -586.4 & -33.8 & -509.8 & -27.3 & -211.8 & -80.0 & -501.0 & -36.7 & h2t \\
5mfg & -1215.8 & -42.3 & -1171.4 & -28.1 & -1195.7 & -15.5 & -1236.5 & -42.2 & -1049.8 & -36.3 & -1214.7 & -40.0 & h2s \\
5mjy & -1527.3 & -79.4 & -1538.3 & -68.7 & -1369.1 & -48.1 & -1482.5 & -58.8 & -1299.2 & -43.8 & -1425.3 & -45.5 & h2s \\
5n4d & -1231.2 & -61.8 & -1206.1 & -62.6 & -1189.6 & -39.4 & -1219.5 & -73.4 & -1034.4 & -35.4 & -1184.8 & -40.9 & s2s \\
5nl1 & -669.8 & -79.2 & -677.0 & -87.6 & -702.1 & -91.6 & -572.6 & -69.0 & -354.6 & -56.7 & -532.3 & -44.3 & s2s \\
5txe & -955.6 & -49.6 & -930.2 & -48.3 & -892.9 & -48.2 & -883.6 & -39.8 & -860.2 & -38.9 & -879.2 & -41.1 & h2t \\
5vt9 & -622.4 & -128.5 & -546.2 & -97.0 & -520.6 & -72.4 & -469.3 & -86.0 & 1.8 & -58.2 & -438.5 & -51.6 & h2s \\
5wkf & 258.4 & -72.2 & 343.2 & -36.4 & 384.2 & -25.3 & 322.4 & -42.8 & 408.0 & -38.8 & 303.2 & -54.4 & s2t \\
5wpl & -352.8 & -99.6 & -380.7 & -91.3 & -284.4 & -59.4 & -216.6 & -69.4 & N/A & N/A & -197.3 & -51.4 & s2s \\
5yis & -416.4 & -99.1 & -291.1 & -34.1 & -263.1 & -57.8 & -265.5 & -50.1 & -149.5 & -50.0 & -302.1 & -56.1 & s2s \\
    \bottomrule
    \end{tabular}\label{tab:appendix_evaluation_all_1}
    \renewcommand{\arraystretch}{1}
\end{table*}

\begin{table*}[t!]
    \scriptsize
    \centering
    \caption{Stability and affinity of the reference peptide, linear peptides designed by baseline methods, and cyclic peptide designed by our method along with the cyclization type.}
    \renewcommand{\arraystretch}{1.2}
    \begin{tabular}{l|rr|rr|rr|rr|rr|rrr}
    \toprule
    \multirow{2}{*}{Target}
    & \multicolumn{2}{c|}{Reference}
    & \multicolumn{2}{c|}{RFDiffusion}
    & \multicolumn{2}{c|}{ProteinGenerator}
    & \multicolumn{2}{c|}{PepFlow}
    & \multicolumn{2}{c|}{PepGLAD}
    & \multicolumn{3}{c}{Our}
    \\
    \cmidrule(lr){2-3} \cmidrule(lr){4-5} \cmidrule(lr){6-7} 
    \cmidrule(lr){8-9} \cmidrule(lr){10-11} \cmidrule(lr){12-14}
    & Stab. & Affi. & Stab. & Affi. & Stab. & Affi. & Stab. & Affi. & Stab. & Affi. & Stab. & Affi. & Type \\
    \midrule
5zw6 & -361.4 & -46.9 & -298.0 & -14.3 & -497.5 & -29.3 & -338.0 & -41.4 & -306.9 & -28.5 & -364.7 & -51.5 & h2s \\
6bli & -1269.1 & -104.4 & -1231.7 & -62.2 & -1231.4 & -46.5 & -1147.9 & -48.4 & N/A & N/A & -1094.2 & -39.5 & h2s \\
6cv1 & -490.1 & -109.1 & -425.7 & -131.3 & -284.3 & -44.2 & -254.6 & -38.5 & N/A & N/A & -234.9 & -36.1 & h2t \\
6di8 & -1850.6 & -73.9 & -1805.2 & -49.3 & -1795.9 & -44.0 & -1783.6 & -52.4 & -1364.2 & -37.5 & -1793.5 & -59.9 & h2t \\
6dtg & -874.6 & -60.3 & -792.4 & -19.5 & -698.6 & -27.9 & -847.4 & -32.4 & -783.4 & -22.8 & -898.9 & -52.4 & h2s \\
6f0h & -665.8 & -80.0 & -499.4 & -57.0 & -544.4 & -46.0 & -529.0 & -32.7 & -280.6 & -54.8 & -539.4 & -38.9 & h2s \\
6f6d & -771.9 & -88.3 & -666.2 & -47.1 & -653.4 & -63.8 & -674.9 & -39.8 & -599.7 & -25.0 & -666.4 & -39.3 & h2t \\
6g68 & -303.2 & -121.6 & -349.0 & -128.1 & -352.1 & -126.1 & -148.8 & -83.8 & N/A & N/A & -15.7 & -57.9 & h2t \\
6ghr & -1342.1 & -56.9 & -1198.8 & -70.7 & -1260.8 & -49.8 & -1212.9 & -30.0 & -666.1 & -43.8 & -1224.1 & -50.6 & h2t \\
6ict & -1323.9 & -87.2 & -1259.5 & -37.0 & -1222.3 & -29.4 & -1142.0 & -26.6 & -662.6 & -36.7 & -1238.7 & -57.0 & h2s \\
6igk & -785.6 & -104.9 & -768.3 & -84.0 & -762.4 & -64.6 & -698.1 & -75.3 & -472.5 & -52.7 & -680.6 & -62.2 & h2t \\
6jbk & -938.6 & -79.7 & -933.7 & -76.5 & -881.2 & -42.2 & -871.9 & -50.2 & -375.9 & -63.7 & -847.5 & -40.5 & h2s \\
6ocp & -754.5 & -51.1 & -672.6 & -15.4 & -565.0 & -31.0 & -716.6 & -30.7 & -552.5 & -22.7 & -676.5 & -34.2 & h2s \\
6om4 & -690.0 & -67.6 & -582.3 & -49.2 & -506.8 & -9.4 & -662.7 & -25.1 & -485.6 & -22.6 & -710.9 & -47.6 & h2s \\
6p02 & -1134.3 & -202.1 & -1136.5 & -171.3 & -1059.6 & -150.6 & -861.6 & -83.6 & -679.6 & -81.1 & -854.0 & -94.5 & h2s \\
6peu & -893.2 & -78.2 & -832.3 & -46.2 & -736.2 & -7.2 & -935.1 & -61.8 & -807.9 & -22.6 & -860.5 & -44.7 & s2s \\
6q5r & -529.4 & -94.9 & -585.9 & -132.9 & -566.2 & -109.0 & -437.5 & -98.9 & N/A & N/A & -401.3 & -60.9 & s2t \\
6qs1 & -917.2 & -45.4 & -847.3 & -65.4 & -871.5 & -36.2 & -893.4 & -38.9 & -787.7 & -23.0 & -903.7 & -52.6 & h2s \\
6r16 & -1330.4 & -102.6 & -1309.5 & -82.5 & -1263.7 & -47.8 & -1196.3 & -50.6 & -440.0 & -64.7 & -1181.7 & -50.5 & h2t \\
6rqx & -1176.7 & -28.1 & -1069.6 & -32.2 & -1078.0 & -19.0 & -1151.5 & -20.3 & -1100.8 & -15.3 & -1207.8 & -37.6 & s2s \\
6rxr & -449.0 & -66.1 & -472.8 & -49.5 & -427.0 & -30.4 & -466.3 & -35.8 & -80.7 & -42.2 & -499.6 & -192.0 & s2t \\
6sa8 & -461.4 & -62.0 & -307.1 & -44.0 & -364.0 & -25.4 & -404.2 & -35.4 & -247.5 & -27.8 & -424.9 & -33.1 & h2s \\
6trw & -1085.1 & -44.4 & -1074.5 & -48.0 & -947.0 & -40.2 & -1093.9 & -36.1 & -544.2 & -37.3 & -1080.4 & -48.9 & h2t \\
6y1a & -393.6 & -202.3 & -358.1 & -191.0 & 39.8 & -49.7 & -153.8 & -84.7 & 264.3 & -68.4 & -142.2 & -69.2 & h2t \\
6zw0 & -589.9 & -105.7 & -576.7 & -115.6 & -405.2 & -18.5 & -458.6 & -61.4 & -294.6 & -38.9 & -426.0 & -50.5 & h2s \\
7atr & -1056.1 & -66.1 & -912.3 & -35.8 & -1036.8 & -33.5 & -991.2 & -37.1 & -978.0 & -33.5 & -999.1 & -49.0 & h2t \\
7brk & -561.5 & -89.4 & -594.4 & -91.7 & -614.1 & -91.1 & -486.8 & -47.7 & -345.7 & -49.7 & -487.3 & -52.1 & h2s \\
7eib & -342.6 & -50.3 & N/A & N/A & -331.3 & -45.8 & -276.4 & -34.4 & -270.9 & -37.7 & -387.1 & -63.7 & h2s \\
7f6h & -238.6 & -50.3 & N/A & N/A & -149.4 & -24.6 & -186.3 & -38.7 & -185.7 & -35.7 & -271.7 & -57.3 & h2s \\
7mhz & -187.5 & -31.9 & N/A & N/A & -74.8 & -27.0 & -148.7 & -29.8 & -48.7 & -34.2 & -217.9 & -59.5 & h2t \\
7okp & -845.2 & -36.8 & N/A & N/A & -721.8 & -20.4 & -793.1 & -27.6 & -522.7 & -17.2 & -857.3 & -36.6 & h2t \\
7owu & -469.1 & -50.8 & N/A & N/A & -454.9 & -39.1 & -464.5 & -23.4 & -293.0 & -27.0 & -487.5 & -62.8 & s2t \\
7q66 & -375.5 & -189.6 & N/A & N/A & -306.4 & -130.5 & -17.6 & -88.6 & 241.7 & -54.4 & -96.6 & -80.0 & h2s \\
7ure & -41.8 & -50.4 & -65.1 & -61.4 & 101.5 & -40.2 & 120.6 & -23.2 & 114.4 & -29.2 & 56.4 & -40.7 & h2s \\
7vb7 & -156.8 & -92.4 & 62.6 & -78.2 & -23.8 & -33.1 & 47.1 & -38.2 & N/A & N/A & 19.1 & -20.6 & h2s \\
7vwo & -647.0 & -93.5 & -519.6 & -58.2 & -435.8 & -41.1 & -464.2 & -44.9 & -125.6 & -72.4 & -523.5 & -54.8 & h2t \\
7wvx & -314.7 & -81.9 & -170.0 & -54.2 & -221.3 & -45.4 & -159.7 & -40.5 & -177.8 & -58.6 & -254.4 & -64.9 & h2t \\
7xxf & -243.6 & -73.7 & -281.3 & -71.4 & -259.1 & -71.7 & -154.1 & -54.7 & -64.0 & -54.4 & -115.9 & -44.2 & h2s \\
7yat & -655.4 & -247.2 & -632.5 & -233.3 & -234.9 & -39.4 & -356.3 & -82.7 & 859.3 & -65.8 & -532.1 & -241.8 & h2s \\
8dgq & -843.5 & -91.9 & -760.5 & -34.4 & -654.9 & -23.8 & -621.5 & -18.1 & -651.9 & -41.0 & -710.6 & -40.4 & h2t \\
    \bottomrule
    \end{tabular}\label{tab:appendix_evaluation_all_part_2}
    \renewcommand{\arraystretch}{1}
\end{table*}

\label{app:limitations}
\section{Limitations and Future Work} 

One limitation is that the generated cyclic peptides may sometimes exhibit invalid conformations, such as inaccurate bond lengths and atomic receptor clashes. While Rosetta-based structure relaxation \citep{chaudhury2010pyrosetta,alford2017rosetta} can refine these structures, it is computationally expensive and slow. Additionally, for evaluation, we are currently unable to introduce self-consistency metrics similar to those in protein design \citep{yim2024improvedmotifscaffoldingse3flow}, as there is no highly accurate cyclic peptide structure prediction or docking model available.

Property-guided sampling can be incorporated during generation to generate chemically and structurally valid cyclic peptides \citep{dhariwal2021diffusion,ho2022classifier}. For example, the process can be conditioned on predefined bond length and angle distributions to sample cyclic peptides with specific shapes. 
Additionally, energy-based sampling \citep{lu2023contrastive,kulyte2024improving} and energy-based preference optimization \citep{zhou2024decompopt,zhou2024stabilizing,zhou2024antigen,cheng2024decomposed} can guide the generation of low-energy, stable conformations. 
Techniques and architectures related to AlphaFold 3 could also be leveraged for more accurate atomic interaction modeling \citep{abramson2024accurate}. For evaluation, we believe it is crucial to validate the cyclic peptides generated by our model in the wet lab, determining their accurate structural conformations and binding modes—an avenue we are actively exploring.

Additionally, we would like to point out that our current method does not explore how to automatically select the best cyclization type for a given receptor, although it can be enumerated. We plan to investigate this aspect in our future work. Other future research includes cyclic ligand peptide design considering flexible protein targets \citep{zhou2025integrating} or dual targets \citep{zhou2024reprogramming}.

\end{document}